\documentclass[11pt,a4paper]{article}
\usepackage[hyperref]{imports/acl2021}

\usepackage{times}
\usepackage{latexsym}

\usepackage{microtype}
\usepackage[utf8]{inputenc} % allow utf-8 input
\usepackage[T1]{fontenc}    % use 8-bit T1 fonts
\usepackage{booktabs}       % professional-quality tables
\usepackage{amsfonts}       % blackboard math symbols
\usepackage{nicefrac}       % compact symbols for 1/2, etc.
\usepackage{microtype}      % microtypography
\usepackage{xcolor}

\usepackage{graphicx}
\usepackage{subfigure}
\usepackage{soul}
\usepackage{amssymb}
\usepackage{bbm}

\usepackage{enumitem}
\usepackage{algorithm}
\usepackage[normalem]{ulem}
\usepackage[noend]{algpseudocode}

\usepackage[colorinlistoftodos,prependcaption,textsize=tiny]{todonotes}

\usepackage{comment}

\usepackage{natbib}

\usepackage{amsmath,amsfonts,bm}
\usepackage{fixltx2e}
\usepackage{multirow}
\usepackage{tabularx}
\usepackage{caption}
\usepackage{wrapfig}
\usepackage[export]{adjustbox}
\usepackage{xspace}
\usepackage{kotex}

\usepackage{caption}
\definecolor{auburn}{rgb}{0.43, 0.21, 0.1}
\definecolor{ao(english)}{rgb}{0.0, 0.5, 0.0}

\newcommand{\red}[1]{\textcolor{red}{#1}}
\newcommand{\blue}[1]{\textcolor{blue}{#1}}

\newcommand{\auburn}[1]{\textcolor{auburn}{#1}}
\newcommand{\green}[1]{\textcolor{ao(english)}{#1}}

\newcommand{\ie}{{\em i.e.,}\xspace}
\newcommand{\eg}{{\em e.g.,}\xspace}
\newcommand{\wrt}{\emph{w.r.t.}\xspace}

\newcommand{\Ni}{({\em i})~}
\newcommand{\Nii}{({\em ii})~}

\def\eqref#1{equation~\ref{#1}}
\def\1{\bm{1}}

\DeclareMathAlphabet{\mathsfit}{\encodingdefault}{\sfdefault}{m}{sl}
\SetMathAlphabet{\mathsfit}{bold}{\encodingdefault}{\sfdefault}{bx}{n}

\def\gD{{\mathcal{D}}}

\def\gH{{\mathcal{H}}}

\def\gL{{\mathcal{L}}}

\def\gN{{\mathcal{N}}}

\def\gX{{\mathcal{X}}}
\def\gY{{\mathcal{Y}}}

\newcommand{\E}{\mathbb{E}}

\newcommand{\multimix}{\textsc{Uxla}}

\DeclareMathOperator*{\argmax}{arg\,max}

\DeclareMathOperator{\real}{\rm I\!R}

\usepackage[nameinlink]{cleveref}
\crefformat{section}{\S#2#1#3} % see manual of cleveref, section 8.2.1
\crefformat{subsection}{\S#2#1#3}
\crefformat{subsubsection}{\S#2#1#3}

\usepackage{enumitem}
\usepackage[normalem]{ulem}

\aclfinalcopy % Uncomment this line for the final submission
\usepackage{cleveref}
\crefformat{section}{\S#2#1#3} % see manual of cleveref, section 8.2.1
\crefformat{subsection}{\S#2#1#3}
\crefformat{subsubsection}{\S#2#1#3}

\title{\multimix: A Robust Unsupervised Data Augmentation Framework for {Zero-Resource} Cross-Lingual NLP}

\author{M Saiful Bari \thanks{\ \ Equal contribution}$^{\ \ \P}$, Tasnim Mohiuddin $^{*\P}$, and Shafiq Joty$ ^{\P\S}$  \\
  $^\P$ Nanyang Technological University, Singapore \\
  $^\S$ Salesforce Research Asia, Singapore \\
  \texttt{\{bari0001@e., mohi0004@e., srjoty@\}ntu.edu.sg } \\}

\date{}

\begin{document}
\maketitle
\begin{abstract}
Transfer learning has yielded state-of-the-art (SoTA) results in many supervised NLP tasks. However,  annotated data for every target task in every target language is rare, especially for low-resource languages. We propose \multimix\, a novel unsupervised data augmentation framework for zero-resource transfer learning scenarios. In particular, \multimix\ aims to solve cross-lingual adaptation problems from a source language task distribution to an unknown target language task distribution, assuming no training label in the target language. At its core, \multimix\ performs simultaneous self-training with data augmentation and unsupervised sample selection. To show its effectiveness, we conduct extensive experiments on three diverse zero-resource cross-lingual transfer tasks. \multimix\ achieves SoTA results in all the tasks, outperforming the baselines by a good margin. With an in-depth framework dissection, we demonstrate the cumulative contributions of different components to its success. %Our code will be released.
%for Named Entity Recognition (NER), Natural Language Inference (NLI) and paraphrase identification on Paraphrase Adversaries from Word Scrambling (PAWS). 
\end{abstract}

\section{Introduction}  \label{sec:intro}

Self-supervised learning in the form of pretrained language models (LM) has been the driving force in developing state-of-the-art NLP systems in recent years. These methods typically follow two basic steps, where a \emph{supervised} task-specific fine-tuning follows a large-scale LM pretraining \cite{radford2019language}. However, getting labeled data for every target task in every target language is difficult, especially for low-resource languages. 

Recently, the \emph{pretrain-finetune} paradigm has also been extended to multi-lingual setups to train effective multi-lingual models that can be used for {zero-shot} cross-lingual transfer. Jointly trained deep multi-lingual LMs like mBERT \citep{BERT} and XLM-R \citep{XLMR} coupled with supervised fine-tuning in the source language have been quite successful in transferring linguistic and task knowledge from one language to another without using any task label in the target language. The joint pretraining with multiple languages allows these models to generalize across languages.

Despite their effectiveness,  recent studies \citep{pires-etal-2019-multilingual,k2020crosslingual} have also highlighted one crucial limiting factor for successful cross-lingual transfer. They all agree that the cross-lingual generalization ability of the model is limited by the (lack of) structural similarity between the source and target languages. For example, for transferring mBERT from English, \citet{k2020crosslingual} report about $23.6\%$ accuracy drop in Hindi (structurally dissimilar) compared to $9\%$ drop in Spanish (structurally similar) in cross-lingual natural language inference (XNLI). {The difficulty level of transfer is further exacerbated if the (dissimilar) target language is low-resourced, as the joint pretraining step may not have seen many instances from this language in the first place. In our experiments (\cref{sec:results}), in cross-lingual NER (XNER), we report F1 reductions of 28.3\% in Urdu and 30.4\% in Burmese for XLM-R, which is trained on a much larger multi-lingual dataset than mBERT.%\todo{put the drops}
}

One attractive way to improve cross-lingual generalization is to perform \emph{data augmentation} \cite{Simard1998}, and train the model  on examples that are similar but different from the labeled data in the source language. Formalized by the Vicinal Risk Minimization (VRM) principle \citep{VRM_NIPS2000}, such data augmentation methods have shown impressive results in  vision \citep{zhang2018mixup,MixMatch}. These methods enlarge the support of the training distribution by generating {new} data points from a \emph{vicinity distribution} around each training example. For images, the vicinity of a training image can be defined by a set of operations like rotation and scaling, or by linear mixtures of features and labels \citep{zhang2018mixup}. However, when it comes to text, such \emph{unsupervised} augmentation methods have rarely been successful. The main reason is that unlike images,  linguistic units  are discrete and a smooth change in their embeddings may not result in a plausible linguistic unit that has similar meanings. %\blue{thus would require corresponding changes in the labels to keep the data consistent.}

In NLP, to the best of our knowledge,  the most successful augmentation method has so far been {back-translation} \citep{backtranslate_sennrich-etal-2016-improving} which paraphrases an input sentence through round-trip translation. However, it requires parallel data to train effective machine translation systems, acquiring which can be more expensive for low-resource languages than annotating the target language data. Furthermore, back-translation is only applicable in a supervised setup and to tasks where it is possible to find the alignments between  the original labeled entities and the back-translated entities, such as in question answering \citep{wei2018fast}. {Other related work includes {contextual augmentation} \cite{kobayashi}, {conditional BERT} \citep{conditional_bert} and AUG-BERT {\cite{Shi2019AUGBERTAE}}. These methods use a constrained augmentation that alters a pretrained LM to a label-conditional LM for a specific task. Since they rely on labels, their application is limited by the availability of enough task labels.}

%  This means these methods update the parameters of the pretrained LM using the labels.

% In this work, we propose \multimix\, a generic data augmentation strategy for improving cross-lingual generalization of multilingual pretrained language models. \multimix uses a pretrained language model to draw virtual input samples (\emph{e.g.}, sentences) from the vicinity distribution of the training examples in the source and target languages. It then performs simultaneous \emph{self-learning} with \emph{tri-training} strategy  \cite{Zhou2005TritrainingEU} to learn an effective cross-lingual model from noisy (pseudo) labels for the target task. We propose novel ways to generate new training samples using a multilingual masked LM \cite{XLMR}, and get reliable task labels by simultaneous multi-lingual co-training.

In this work, we propose \multimix, a robust \textbf{u}nsupervised \textbf{cross}-\textbf{l}ingual \textbf{a}ugmentation framework for improving cross-lingual generalization of multilingual LMs. \multimix\ augments data from the unlabeled training examples in the target language as well as from the virtual input samples generated from the vicinity distribution of the source and target language sentences. With the augmented data, it performs simultaneous \emph{self-learning} with an effective \emph{distillation strategy} to learn a strongly adapted cross-lingual model from noisy (pseudo) labels for the target language task. We propose novel ways to generate virtual sentences using a multilingual masked LM \citep{XLMR}, and get reliable task labels by simultaneous multilingual co-training. {This co-training employs a two-stage co-distillation process to ensure robust transfer to dissimilar and/or low-resource languages.}

%\blue{A two-stage co-distillation process is embedded inside the co-training to ensure robust transfer to dissimilar and/or low-resource languages.}  

%\red{Our augmented data come from the unlabeled monolingual training examples of the target language as well as from the virtual input samples (\emph{e.g.}, sentences) from the vicinity distribution of the unlabeled examples of the source and target languages.} 

We validate the effectiveness {and robustness} of \multimix\ by performing extensive experiments on {three} diverse zero-resource cross-lingual transfer tasks--{XNER, XNLI, and PAWS-X}, which posit different sets of challenges, and across many (14 in total) language pairs comprising languages that are similar/dissimilar/low-resourced. \multimix\ yields impressive results on XNER, setting SoTA in all tested languages outperforming the baselines by a good margin. The relative gains for \multimix\ are particularly higher for structurally dissimilar and/or low-resource languages: 28.54\%, 16.05\%, and 9.25\% absolute improvements for Urdu, Burmese, and Arabic, respectively. For XNLI, with only 5\% labeled data in the source, it gets comparable results to the baseline that uses all the labeled data, and surpasses the standard baseline by 2.55\% on average when it uses all the labeled data in the source. {We also have similar findings in PAWS-X.} We provide a comprehensive analysis of the factors that contribute to \multimix’s performance. We open-source our framework at \href{https://ntunlpsg.github.io/project/uxla/}{https://ntunlpsg.github.io/project/uxla/} .

\begin{figure*}[t!]
  \centering
\scalebox{.85}{
%https://drive.google.com/file/d/1NmAnLY8L_pCZAvsuUfEZz3BQf3wwO4PD/view?usp=sharing
%https://drive.google.com/file/d/1wM3QoZPIACbyDZ00l1XpdTm4KxT6ivMG/view?usp=sharing
  \includegraphics[width=1\linewidth,trim=.1 .1 .1 .1,clip]{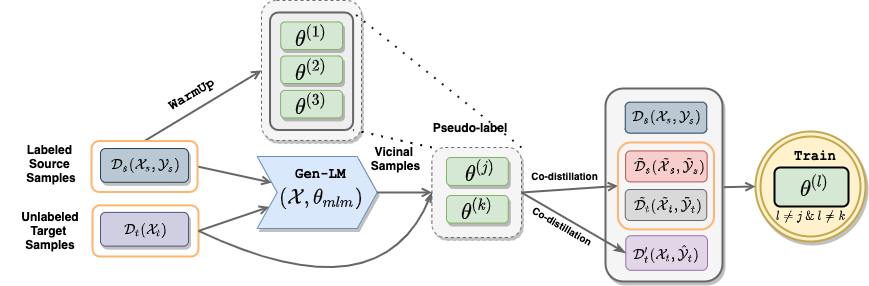}}
\caption{\small Training flow of \multimix. After training the base task models $\theta^{(1)}$, $\theta^{(2)}$, and $\theta^{(3)}$ on source labeled data $\gD_s$ (\textbf{WarmUp}), we use two of them ($\theta^{(j)}$, $\theta^{(k)}$) to \textbf{pseudo-label} and \textbf{co-distill} the  unlabeled target language data ($\gD_t^{\prime}$). A pretrained LM (\textbf{Gen-LM}) is used to generate new vicinal samples for both source and target languages, which are also pseudo-labeled and co-distilled using the two task models ($\theta^{(j)}$, $\theta^{(k)}$) to generate $\tilde{\gD}_s$ and $\tilde{\gD}_t$. The third model $\theta^{(l)}$ is then progressively trained on these datasets: $\{\gD_s, \gD'_t\}$ in epoch 1, $\tilde{\gD}_t$ in epoch 2, and all in epoch 3.}  
\label{fig:proposed-model}
\end{figure*}

\section{\multimix\ Framework}  \label{sec:method}
% \vspace{-0.5em}
%\red{DISTIL not defined}\\
%\red{zero-resource vs. zero-shot -> will describe in setup or appendix} \\
%\red{tri-training paper cite}\\
%In this section, we present \multimix\, our proposed data augmentation framework for zero-resource cross-lingual learning. \multimix\ augments new samples in both source and target languages and performs unsupervised sample selection with self-training. Figure \ref{fig:proposed-model} gives an overview of the method. 

%To learn a strongly adapted cross-lingual model from noisy labels and to exclude the outliers for the target task, our framework uses a distillation strategy.

%To avoid \emph{confirmation bias} with self-learning where the model accumulates its own errors, we simultaneously train three models to generate `virtual' training data through data augmentation and filter potential noises for each other through epoch level co-teaching. In our case, each of these models is an instance of an mBERT \cite{BERT}.  

% A pre-trained language model (LM) (parameterized by $\theta_{lm}$) is used to define the vicinity distribution around each source training example and each {\textit{selected}} target example. The newly generated training samples (\emph{e.g.},\ sentences without labels) are then labeled with the corresponding model(s) to create the associated `\textit{pseudo}' labels. 

%In this way, in each epoch, each of the three models is trained with the augmented source and/or target language data filtered by the other two peer models with the proposed \textit{distillation} method.

While recent cross-lingual transfer learning efforts have relied almost exclusively on {multi-lingual} pretraining and zero-shot transfer of a fine-tuned source model, we believe there is a great potential for more elaborate methods that can leverage the unlabeled data better. Motivated by this, we present \multimix, our unsupervised data augmentation framework for zero-resource cross-lingual task adaptation. Figure \ref{fig:proposed-model} gives an overview of \multimix.

%Figure \ref{fig:proposed-model} gives an overview of \multimix\. 

 Let $\gD_s = (\gX_s, \gY_s)$ and $\gD_t = (\gX_t)$ denote the training data for a source language $s$ and a target language $t$, respectively. \multimix\ augments data from various origins at different stages of training. In the initial stage {(epoch 1)}, it uses the {augmented} training samples from the target language ($\gD_t^{\prime}$) along with the original source ($\gD_s$). In later stages {(epoch 2-3)}, it uses vicinal sentences generated from the vicinity distribution of source and target examples: $\vartheta(\Tilde{x}_n^s|{x}_n^s)$ and $\vartheta(\Tilde{x}_n^t|{x}_n^t)$, where ${x}_n^s \sim \gX_s$ and ${x}_n^t \sim \gX_t$. It performs \emph{self-training} on the augmented data to acquire the corresponding {pseudo} labels. To avoid \emph{confirmation bias} with self-training where the model accumulates its own errors, it simultaneously trains three task models  to generate \emph{virtual} training data through data augmentation and filtering of potential label noises via {multi-epoch} \emph{co-teaching} \citep{tritraining}.
 
In each epoch, the co-teaching process first performs \emph{co-distillation}, where two peer task models are used to select ``reliable'' training examples to train the third model. The selected samples with pseudo labels are then added to the target task model's training data by taking the {agreement} from the other two models, a process we refer to as \emph{co-guessing}. The co-distillation and co-guessing mechanism ensure robustness of \multimix\ to out-of-domain distributions that can occur in a multilingual setup, \eg\ due to a structurally dissimilar and/or low-resource target language. \Cref{alg:XLA} gives a pseudocode of the overall training method. Each of the \emph{task} models in \multimix\ is an instance of XLM-R fine-tuned on the source language task (e.g., English NER), whereas the pretrained masked LM parameterized by $\theta_{\text{mlm}}$ (\ie\ before fine-tuning) is used to define the \emph{vicinity} distribution $\vartheta(\Tilde{x}_n|{x}_n, \theta_{\text{mlm}})$ around each selected example ${x}_n$. In the following, we describe the steps in \Cref{alg:XLA}.

\begin{algorithm*}[t!]
\footnotesize
\caption{\multimix: a robust unsupervised data augmentation framework for cross-lingual NLP}
\label{alg:XLA}
 \textbf{Input:} \textit{source} (s) and \textit{target} (t) language datasets: $\gD_s = (\gX_s, \gY_s), \gD_t = (\gX_t)$; task models: $\theta^{(1)}, \theta^{(2)}, \theta^{(3)}$, pre-trained masked LM {$\theta_{\text{mlm}}$}, mask ratio $P$, diversification factor $\delta$, {sampling factor $\alpha$}, and distillation factor $\eta$\\
 \textbf{Output:} models trained on augmented data
    % \hspace*{\algorithmicindent} 
\begin{algorithmic}[1]
    \State $\theta^{(1)}, \theta^{(2)}, \theta^{(3)} = \textsc{WarmUp}(\gD_s, \theta^{(1)}, \theta^{(2)}, \theta^{(3)})$ \algorithmiccomment{warm up with \blue{conf. penalty}.} 
    
    %\algorithmiccomment{standard src. language training (with conf. penalty $\mathcal{H}$)}
    
    \For{$e \in [1:3]$} \algorithmiccomment{$e$ denotes epoch.}
    \For{$k \in \{1, 2, 3 \}$}
    \State $\gX_t^{(k)}, \gY_t^{(k)} = \textsc{Distil} (\gX_t, \eta_e, \theta^{(k)})$ \algorithmiccomment{infer and select tgt training data for augmentation.}
    % \State $\gX_t^{(2)}, \gY_t^{(2)} = \textsc{Distill} (\gD_t, K, \theta^{(2)})$ 
    % \State $\gX_t^{(3)}, \gY_t^{(3)} = \textsc{Distill} (\gD_t, K, \theta^{(3)})$ 

    \For{$j \in \{1, 2, 3 \}$}
    \If{$k == j$} 
    Continue
    \EndIf
    
    \State \texttt{/* \blue{source language data augmentation} */}

    \State $\tilde{\gX}_s = \textsc{Gen-LM}(\gX_s,\theta_{\text{mlm}}, P, \delta)$ \algorithmiccomment{vicinal example generation.}  %\algorithmiccomment{generate for source data augmentation.}
    \State $\gX_s^{(k)}, \gY_s^{(k)} = \textsc{Distil}(\tilde{\gX}_s, \eta_e, \theta^{(k)})$; \hspace{1em} $\gX_s^{(j)}, \gY_s^{(j)} = \textsc{Distil}(\tilde{\gX}_s, \eta_e,  \theta^{(j)})$
%    \State $\gX_s^{(j)}, \gY_s^{(j)} = \textsc{Distil}(\tilde{\gX}_s, \eta_e,  \theta^{(j)})$
    \State $\tilde{\gD}_s = \textsc{Agreement} \big( \gD_s^{(k)}=( \gX_s^{(k)}, \gY_s^{(k)} ), \gD_s^{(j)} = ( \gX_s^{(j)}, \gY_s^{(j)}) \big)$
    % \algorithmiccomment{augment source data.}
    
    \State \texttt{/* \green{target language data augmentation (no vicinity)} */}
    \State $\gX_t^{(j)}, \gY_t^{(j)} = \textsc{Distil} (\gX_t, \eta_e, \theta^{(j)})$ \State $\gD_t^{\prime} = \textsc{Agreement} \big( \gD_t^{(k)}=( \gX_t^{(k)}, \gY_t^{(k)} ), \gD_t^{(j)} = ( \gX_t^{(j)}, \gY_t^{(j)}) \big)$\algorithmiccomment{see line 4}
    % \State $\gX_t^{'} = \gX_t^{(k)} \cap \gX_t^{(j)}$
    % \State $\gX_t^{(k)}, \gY_t^{(k)} = \textsc{Label}(\gX_t^{'},\theta^{(k)})$ 
    % \State $\gX_t^{(j)}, \gY_t^{(j)} = \textsc{Label}(\gX_t^{'},\theta^{(j)})$
    %\algorithmiccomment{select common target training data.}
    \State \texttt{/* \red{target language data augmentation} */}
    \State $\tilde{\gX}_t = \textsc{Gen-LM}(\gX_t,\theta{\text{mlm}}, P, \delta)$ \algorithmiccomment{vicinal example generation.} %\algorithmiccomment{generate for target data augmentation.}
    \State $\gX_t^{(k)}, \gY_t^{(k)} = \textsc{Distil}(\tilde{\gX}_t, \eta_e, \theta^{(k)})$; \hspace{1em}  $\gX_t^{(j)}, \gY_t^{(j)} = \textsc{Distil}(\tilde{\gX}_t, \eta_e, \theta^{(j)})$
%    \State $\gX_t^{(j)}, \gY_t^{(j)} = \textsc{Distil}(\tilde{\gX}_t, \eta_e, \theta^{(j)})$
    \State $\tilde{\gD}_t = \textsc{Agreement} \big( \gD_t^{(k)}=( \gX_t^{(k)}, \gY_t^{(k)} ), \gD_t^{(j)} = ( \gX_t^{(j)}, \gY_t^{(j)}) \big)$ %\algorithmiccomment{augment target data.}
   
    \State \texttt{/* {\auburn{train new models on augmented data}} */}
    
    \For{$l \in \{1, 2, 3 \}$}
    \If{$l \ne j \text{ and } l \ne k $}
    \State \hspace{-0.5em} {with sampling factor $\alpha$, train $\theta^{(l)}$ on $\gD$,} \algorithmiccomment{train \blue{progressively}}
    \State \hspace{-0.5em}  where $\gD = \{ \gD_s \mathbbm{1} (e \in \{1,3\})  \cup \gD_t^{\prime} \mathbbm{1} (e \in \{1,3\})  \cup \tilde{\gD}_s \mathbbm{1}(e = 3) \cup \tilde{\gD}_t\mathbbm{1} (e \in \{2,3\} ) \}$

    %$\gD $ %\algorithmiccomment{Train a new model.}
%    \State where $\gD = \{ \gD_s\mathbf{1}_{\{1,3\}}(e)  \cup \gD_t^{\prime}\mathbf{1}_{\{1,3\}}(e) \cup \gD_s^{\text{aug}}\mathbf{1}_{\{3\}}(e) \cup \gD_t^{\text{aug}}\mathbf{1}_{\{2,3\}}(e) \}$
%    \State and $e = epoch$

    \EndIf
    \EndFor
    \EndFor
    \EndFor
    \EndFor  
\State Return $\{\theta^{(1)}, \theta^{(2)}, \theta^{(3)}\}$
\end{algorithmic}
% \vspace{em}
% \vspace{-.3em}
\end{algorithm*}

\subsection{Warm-up: Training Task Models}
\label{subsec:conf-pen}
%\vspace{-.5em}

We first train three instances of the XLM-R model ($\theta^{(1)}, \theta^{(2)}, \theta^{(3)}$) with an additional task-specific linear layer on the source language (English) labeled data. Each model has the same architecture (XLM-R large) but is initialized with different random seeds. For token-level prediction tasks (\eg\, NER), the \textit{token-level} representations are fed into the classification layer, whereas for sentence-level tasks (\eg\, XNLI), the \texttt{[CLS]} representation is used as input to the classification layer. %For each task, we fine-tune all the model parameters end-to-end. %We select the best model based on the validation set accuracy.  

% \vspace{-0.5em}
\paragraph{Training with confidence penalty} 

{Our goal is to train the task models so that they can be used reliably for self-training on a target language that is potentially dissimilar and low-resourced. In such situations, an overly confident (overfitted) model may produce more noisy pseudo labels, and the noise will then accumulate as the  training progresses. Overly confident predictions  may also impose difficulties on our distillation methods (\cref{subsec:distil})} in isolating good samples from noisy ones. However, training with the standard cross-entropy (CE) loss may result in overfitted models that produce overly confident predictions (low entropy), especially when the class distribution is not balanced. We address this by adding a negative entropy term $- \gH$  to the CE loss as follows.

\vspace{-1em}
\begin{equation}
\small 
    \gL(\theta) = \sum_{c=1}^C \Big[ - \underbrace{y^c \log p_{\theta}^c(\mathbf{x})}_\text{CE} + \underbrace {p_{\theta}^c(\mathbf{x}) \log p_{\theta}^c(\mathbf{x})}_{- \gH}  \Big]
\normalsize
\end{equation}
\noindent where $\mathbf{x}$ is the representation that goes to the output layer, and $y^c$ and $p_\theta^c(\mathbf{x})$ are respectively the  ground truth label and model predictions with respect to class $c$. Such regularizer of output distribution has been shown to be effective for training large models \citep{Pereyra17}. We also report significant gains with confidence penalty in \cref{sec:eval}. Appendix B shows visualizations on why confidence penalty is helpful for distillation.  

%We train the task models using the cross-entropy (CE) loss. However, maximum likelihood training with the standard CE loss may result in overfitted models that produce overly confident predictions (low entropy), especially when the class distribution is not balanced. This may impose difficulties on our sample selection methods in isolating good samples from noisy ones \cite{Li2020DivideMix}. Following \cite{Pereyra17}, we address this issue by adding a negative entropy term $- \gH$  to the CE loss.

%\begin{itemize}
%    \item Validation is done with Source language (english).
%    \item Confident Penalty can be added for avoiding model overfit to asymmetric label noise (see divide mix page 4 last paragraph, https://openreview.net/pdf?id=HJgExaVtwr)
%    \item During warmup state 3 identical model, M is trained with 3 different seed.
%\end{itemize}

%separate correct samples from incorrect ones.   

%For each of the sample we calculate a pseudo loss from the prediction of the respective model. Later, we train a GMM to distinguish confident and noisy samples. See divide mix paper, page 3 last paragraph.

% https://openreview.net/pdf?id=HJgExaVtwr
 
% \textcolor{red}{It would be nice to write some eq. for GMM if we have time.}
 
%\subsection{Language Model as the Vicinity Model} \label{sec:LM}

% \vspace{-.5em}
\subsection{Sentence Augmentation} \label{sec:vicinity}
%\vspace{-.5em}

Our augmentated sentences come from two different sources: the \emph{original} target language samples $\gX_t$, and the \emph{virtual} samples generated  from the vicinity distribution of the source and target samples: $\vartheta(\Tilde{x}_n^s|{x}_n^s, \theta_{\text{mlm}})$ and $\vartheta(\Tilde{x}_n^t|{x}_n^t,\theta_{\text{mlm}})$ with ${x}_n^s \sim \gX_s$ and ${x}_n^t \sim \gX_t$. It has been shown that contextual LMs pretrained on large-scale datasets capture useful linguistic features and can be used to generate fluent grammatical texts \citep{hewitt-manning-2019-structural}. We use XLM-R masked LM \citep{XLMR} as our vicinity model $\theta_{\text{mlm}}$, which is trained on massive multilingual {corpora} (2.5 TB of Common-Crawl data in 100 languages). The vicinity model is a disjoint pretrained entity whose parameters are not trained on any task objective. 

%we use XLM-R masked LM ($\theta_{{lm}}$) as a disjoint entity working as a vicinity model whose weights are not trained on any task-specific objective.

% to generate additional samples around each \textit{selected sample}. 

In order to generate samples around each \emph{selected} example, we first randomly choose $P\%$ of the input tokens. Then we successively (one at a time) mask one of the chosen tokens and ask XLM-R masked LM to predict a token in that masked position, \ie\ compute $\vartheta(\Tilde{x}_m |x, \theta_{\text{mlm}})$ with $m$ being the index of the masked token. For a specific mask, we sample $S$ candidate words from the output distribution, and generate novel sentences by following one of the two alternative approaches.

\textbf{(i) Successive max~~} In this approach, we take the most probable output token ($S=1$) at each prediction step, $o_m^* = \argmax_{o} \vartheta(\Tilde{x}_m=o |x, \theta_{\text{mlm}})$. A new sentence is constructed by $P\%$ newly generated tokens. We generate $\delta$ (diversification factor) virtual samples for each original example $x$, by randomly masking $P\%$ tokens each time. %Here, $\delta$ is the .

\textbf{(ii) Successive cross~~} In this approach, we divide each original (multi-sentence) sample $x$  into two parts and use successive max to create two sets of augmented samples of size $\delta_1$ and  $\delta_2$, respectively. We then take the cross of these two sets to generate $\delta_1 \times \delta_2$ augmented samples.

Augmentation of sentences through \textit{successive max} or \emph{cross} is carried out within the $\textsc{Gen-LM}$ (generate via LM) module in \Cref{alg:XLA}. For tasks involving a single sequence (\eg\ XNER), we directly use successive max. Pairwise tasks like XNLI and PAWS-X have pairwise dependencies: dependencies between a premise and a hypothesis in XNLI or  dependencies between a sentence and its possible paraphrase in PAWS-X. To model such dependencies, we use successive cross, which uses cross-product of two successive max applied independently to each component.

\subsection{Co-labeling through Co-distillation}
\label{subsec:distil}
%\vspace{-.2em}

Due to discrete nature of texts, VRM based augmentation methods that are successful for images such as MixMatch \citep{MixMatch} that generates new samples and their labels as simple {linear interpolation}, have not been successful in NLP. The meaning of a sentence can change entirely even with minor variations in the original sentence. For example, consider the following example generated by our vicinity model.

\vspace{0.6em}
\small 
\noindent \textbf{Original:~} \textit{\textcolor{red}{EU} rejects German call to boycott british lamb.}

\noindent \textbf{Masked:~}   \textit{\textcolor{red}{<mask>} rejects german call to boycott british lamb.}
\noindent \textbf{\textsc{XLM-R}:~} \textit{\textcolor{red}{Trump} rejects german call to boycott  british lamb.}
\normalsize
\vspace{0.4em}

\begin{comment}

%\vspace{-0.5em}
\begin{itemize}[leftmargin=*,itemsep=-0.2em,label={}]
%\centering
\footnotesize
%    \item \makebox[2.5cm]{\textbf{Original text:}\hfill}  \textit{\textcolor{red}{EU} rejects German call to boycott british lamb.}
    \item \makebox[2.5cm]{\textbf{Masked text:}\hfill}   \textit{\textcolor{red}{<mask>} rejects german call to boycott british lamb.}
    \item \makebox[2.5cm]{\textbf{MLM prediction:}\hfill}   \textit{\textcolor{red}{Trump} rejects german call to boycott  british lamb.}
\end{itemize}
%\vspace{-0.5em}

\end{comment}

%Traditional VRM based data augmentation methods assume that the samples generated by the vicinity model share the same class  so that the same class labels can be used for the newly generated data \citep{VRM_NIPS2000}. This approach does not consider the vicinity relation across examples of different classes. Recent methods  relax this assumption and generate new samples and their labels as simple \textit{linear interpolations} \citep{MixMatch}. 

%However, due to the discrete nature of texts, such linear interpolation methods have not been successful so far in NLP. 

\noindent Here, EU is an \emph{Organization} whereas the newly predicted word \emph{Trump} is a \emph{Person} (different name type). Therefore, we need to relabel the augmented sentences no matter whether the original sentence has labels (source) or not (target). However, the relabeling process can induce noise, especially for dissimilar/low-resource languages, since the {base} task model may not be adapted fully in the early training stages. We propose a 2-stage {sample distillation process} to filter out noisy augmented data.%, as we describe below.

\paragraph{Stage 1: Distillation by single-model} 

The first stage of distillation involves predictions from a single model for which we propose two alternatives:

\Ni \textit{Distillation by model confidence:~} In this approach, we select samples based on the model's prediction confidence. This method is similar in spirit to the selection method proposed by \citet{ruder-plank-2018-strong}. For sentence-level tasks (\eg\, XNLI), the model produces a single class distribution for each training example. In this case, the model's confidence is computed by $p^* = \max_{c \in \{1 \ldots C\}} p_\theta^c(\mathbf{x})$. 
%\vspace{-1em}
%\begin{equation}
%    \hat{p} = \max_{c \in \{1 \ldots C\}} p_\theta^c(\mathbf{x})
%\end{equation}
%\vspace{-1em}
For token-level sequence labeling tasks (\eg\, NER), the model's confidence is computed by: $p^* = \frac{1}{T} \sum_{t=1}^T \big\{ \max_{c \in \{1 \ldots C\}} p_\theta^c(\mathbf{x}_t) \big\}$, 
%\begin{equation}
%\vspace{-.5em}
%    \hat{p} = \frac{1}{T} \big\{ \max_{c \in \{1 \ldots C\}} p_\theta^c(\mathbf{x}_t) \big\}_{t=1}^T
%\vspace{-.5em}
%\end{equation}
where $T$ is the length of the sequence. The distillation is then done by selecting the top $\eta\%$ samples with the highest confidence scores.%, where $\eta$ is a hyperparameter. 

%If the sample only predicts one probability distribution (for xnli only one logit vector is available) then confidence is measure by $\max( P(x_i)_{i\in \{Class_0, Class_1.. Class_n\}} )$. 

%For sequence labeling problem confidence is measured by $avg ( \max( P(x_i)_{i\in \{Class_0, Class_1.. Class_n\}} ))_{tok_{i=0,1,...,k}}$

%One limitation of the confidence-based distillation is that it does not consider task-specific information. Apart from classifier confidence, there could be other important features that can distinguish a good sample from a noisy one. For example, for sequence labeling, \emph{sequence length} can be an important feature as the models tend to make more mistakes (hence noisy) for longer sequences. One might also want to consider other features like \emph{fluency}, which can be estimated by a pre-trained conditional LM. In the following, we introduce a clustering-based method that can consider these additional features to separate good samples from bad ones.  

%\vspace{-.3em}
\Nii \textit{Sample distillation by clustering:~} {We propose this method based on the finding that large neural models tend to learn good samples faster than noisy ones, leading to a lower loss for good samples and higher loss for noisy ones \citep{NIPS2018Robust,ArazoUnsup2019}. We use a 1d two-component Gaussian Mixture Model (GMM) to model \textit{per-sample loss distribution} and cluster the samples based on their \emph{goodness}. GMMs provide flexibility in modeling the sharpness of a distribution and can be easily fit using \textit{Expectation-Maximization} (EM) (See more on Appendix C). The loss is computed based on the pseudo labels predicted by the model.} For each sample $\mathbf{x}$, its \emph{goodness} probability is the posterior probability $p(z=g|\mathbf{x}, \theta_{\text{GMM}})$, where $g$ is the component with smaller mean loss. Here, distillation hyperparameter $\eta$ is the posterior probability threshold based on which samples are selected.

%Here our goal is to cluster the samples based on their \emph{goodness}. 

%It has been shown in computer vision that deep models tend to learn good samples faster than noisy ones, leading to a lower loss for good samples and higher loss for noisy ones \cite{NIPS2018Robust,Arpit2017}.  

%Inspired by \cite{ArazoUnsup2019,Li2020DivideMix}, we propose to model \textit{per-sample loss distribution} with a mixture model, which we fit using an \textit{Expectation-Maximization} (EM) algorithm. 

%However, contrary to those approaches which use actual (supervised) labels, we use the model predicted pseudo labels to compute the loss for the samples. 

%We use a 1d two-component Gaussian Mixture Model (GMM) due to its flexibility in modeling the sharpness of a distribution \cite{Li2020DivideMix}. 

% \vspace{-.5em}
\paragraph{Stage 2: Distillation by model agreement} In the second stage of distillation, we select samples by taking the agreement (co-guess) of two different peer models $\theta^{(j)}$ and $\theta^{(k)}$ to train the third $\theta^{(l)}$. Formally,
% \vspace{-0.5em}
% \begin{align}
$\textsc{Agreement} \big( \gD^{(k)}, \gD^{(j)}) = \{(\gX^{(k)}, \gY^{(k)}): \gY^{(k)} = \gY^{(j)}\} \ \ \  s.t. ~~ k \neq j$     
% \end{align}

%\blue{We further distil the \textit{primary distilled} augmented data by the agreement (co-guess) of two different peer models ($\theta^{(j)}$ and $\theta^{(k)}$) to train the other model ($\theta^{(l)}$ ).  }
%The generated samples with the predicted labels are then added to the training data for the target model $\theta^{(i)}$  by taking the agreements (co-guess) from the two models. 
%The \textsc{Agreement} method is defined as follows:

%\vspace{-0.5em}
\subsection{Data Samples Manipulation}
\label{data-sample-manipulation}
%\vspace{-.5em}

\multimix\ uses multi-epoch co-teaching. It uses $\gD_s$ and $\gD'_t$ in the first epoch. In epoch 2, it uses $\tilde{\gD}_t$ (target virtual), and finally it uses all the four datasets - $\gD_s$, $\gD'_t$, $\tilde{\gD}_t$, and $\tilde{\gD}_s$ (line 22 in  \Cref{alg:XLA}). The datasets used at different stages can be of different sizes.
%In the final step, \multimix\ trains a model $\theta^{(l)}$ with a combination from four different datasets -- $\gD_s$, $\gD_t$, $\gD_s^{aug}$ and $\gD_t^{aug}$, which can be of different sizes. 
For example, the number of augmented samples in $\tilde{\gD}_s$ and $\tilde{\gD}_t$ grow polynomially with the \emph{successive cross} masking method. Also, the \textit{co-distillation} produces sample sets of variable sizes. To ensure that our model does not overfit on one particular dataset, we employ a balanced sampling strategy. For $N$ number of datasets $\{\gD_i\}_{i=1}^{N}$ with probabilities, $\{p_i\}_{i=1}^{N}$, we define the following multinomial distribution to sample from:% training examples
%\vspace{-0.5em}
\begin{equation}
\small 
p_i = \frac{f_i^{\alpha}}{\sum_{j=1}^{N} f_j^{\alpha}} \text{, where } 
f_i = \frac{n_i}{\sum_{j=1}^{N} n_j} 
\normalsize
\end{equation}
% \vspace{-0.3em}
{\noindent where $\alpha$ is the sampling factor and $n_i$ is the total number of samples in the $i^{th}$ dataset. By tweaking $\alpha$, we can control how many samples a dataset can provide in the mix.}

%{ the original labeled source $\gD_s$, pseudo labeled target $\gD_t$, virtual input samples \Niii $\gD_s^{aug}$ and \Niv $\gD_t^{aug}$ generated from  $\gD_s$ and $\gD_t$ respectively using vicinity model ($\theta_{lm}$). These datasets can be of different sizes. 

%  We define a  from the different data distribution.  they are defined as:

%the augmented data from the virtual input samples generated from the target samples' vicinity distribution, $\gD_t^{aug}$. In the final epoch, we use 

%\red{this section needs to be generic}

%penalize large datasets so that 

%to extract less number of batch from them.

\section{Experiments} \label{sec:eval}

% \vspace{-.3em}
We consider {three} tasks in the \emph{zero-resource} cross-lingual transfer setting. We assume labeled training data  only in English, and transfer the trained model to a target language. For all  experiments, we report the \emph{mean score} of the three models that use different seeds.

% \vspace{-.5em}
\subsection{Tasks \& Settings}
\label{subsec:settings}
%\vspace{-.5em}

\paragraph{XNER:} 
%As a sequence labeling task, XNER evaluates the model's capability to learn task-specific contextual representations that depend on language structure. 
We use the standard CoNLL datasets \citep{Sang2002IntroductionTT,Sang2003IntroductionTT} for English (en), German (de), Spanish (es) and Dutch (nl). We also evaluate on Finnish (fi) and Arabic (ar) datasets collected from \citet{bari19}. Note that Arabic is structurally different from English, and Finnish is from a different language family. To show how the models perform on extremely low-resource languages, we experiment with three structurally different languages from WikiANN \citep{pan-etal-2017-cross} of different (unlabeled) training data sizes: Urdu (ur-20k training samples), Bengali (bn-10K samples), and Burmese (my-100 samples).

\begin{table*}[t]
    \begin{center}
    \scalebox{.7}{
        \begin{tabular}{lcccccc}
        \toprule 
        \textbf{Model} & \textbf{en} & \textbf{es} & \textbf{nl} & \textbf{de} & \textbf{ar} & \textbf{fi}\\
         \midrule
        \multicolumn{7}{c}{\textbf{Supervised Results}} \\
        \midrule
        LSTM-CRF \citep{bari19}  & 89.77  & 84.71  & 85.16   & 78.14  & 75.49  & 84.21  \\ %
        XLM-R \citep{XLMR}  & 92.92 & 89.72 & 92.53 & 85.81 & -- & --\\ %\citep{XLMR}
        {XLM-R (our imp.)}  & 92.9 & 89.2 & 92.9 & 86.2 & 86.8 & 92.4\\ %\citep{XLMR}
        \midrule
        \multicolumn{7}{c}{\textbf{Zero-Resource Baseline}} \\
        \midrule
        %fastText-bi-LSTM-CRF \cite{bari19} & 88.98 $\pm$ 0.25 & x & x & x & x & x \\
        %(Char+fastText)bi-LSTM-CRF \cite{bari19} & 89.92 $\pm$  0.15 & 26.76 $\pm$ 1.45 & 20.94 $\pm$ 0.74 & 8.34 $\pm$ 1.43 & x & 22.44 $\pm$ 2.23 \\
        %\midrule
        %BERT-base-cased & 91.21 $\pm$ 0.18 & 52.88 $\pm$ 1.33 & 29.16 $\pm$ 3.30 & 44.41  $\pm$ 2.36 & x & 30.18 $\pm$ 1.93\\
        % \midrule
        % mBERT-uncased & 90.82 $\pm$ 0.20 & 67.79 $\pm$ 0.75 & 72.20 $\pm$ 0.53 &  73.30 $\pm$ 0.28 & 47.62 $\pm$ 1.41 & 62.98 $\pm$ 0.47\\
        % \citet{MAN_MOE} & - & 73.5  & 72.4 &  56.0 & - & - \\
        
        %\citet{xBERT} & - & 74.96  & 77.57 & 69.56 & - & -\\
        
        %\citet{Keung_2019} & - & 75.0 & 77.6 & 71.9 & - & -\\
        
        %\citet{pires-etal-2019-multilingual} & - & 73.59 &  77.36 & 69.74 & - & -\\
        
        %\citet{XLMR} & - & 78.64 & 80.80  & 71.40 & - & -\\
        
        %\citet{bari19} & - & 75.93 $\pm$ 0.81  & 74.61 $\pm$ 1.24 & 65.24 $\pm$ 0.56 & 36.91 $\pm$ 2.74 & 53.77  $\pm$ 1.54\\
        
        mBERT$_\text{cased}$ (our imp.) & 91.13  & 74.76  & 79.58  &  70.99  & 45.48 & 65.95 \\
        
        XLM-R (our imp.) & 92.23  & 79.29  & 80.87  & 73.40  & 49.04  & 75.57 \\
        
        XLM-R (ensemble) & 92.76 & 80.62  & 81.46  & 75.40  & 52.30  & 76.85 \\
        
        \midrule
        \multicolumn{7}{c}{\textbf{Our Method}} \\
        \midrule
        mBERT$_\text{cased}$ +con-penalty & 90.81 & 75.06  & 79.26 &  72.31 & 47.03 & 66.72 \\
        XLM-R+con-penalty & 92.49 & 80.45 & 81.07 & 73.76 & 49.94 & 76.05\\
        %\multimix\-TopConf &  -  & \textbf{78.40} $\pm$ \textbf{0.56} & 80.42 $\pm$ 0.26 & 73.98  $\pm$ 0.45 & \textbf{53.53} $\pm$ \textbf{0.84} &  \textbf{69.75}	 $\pm$ \textbf{0.63} \\
        %\multimix\-GMM &  -  & 78.36 $\pm$ 0.24 & \textbf{81.51} $\pm$ \textbf{0.63} & \textbf{74.42} $\pm$ \textbf{0.34} & 52.8	 $\pm$ 0.31 &  69.78		 $\pm$ 0.55 \\
        %\midrule
        %\multimix\ & -- & 82.34 $\pm$ 0.42 & 83.51 $\pm$ 0.49  & 78.69 $\pm$ 0.86& 54.05 $\pm$ 0.48 & 79.44 $\pm$ 0.25 \\
        %\multimix\-with-model-agreement 
        \multimix\ & -- & {83.05} & {85.21}  & {80.33} & {57.35} & {79.75} \\
        
        \multimix\ (ensemble) & -- & \textbf{83.24} & \textbf{85.32}  & \textbf{80.99} & \textbf{58.29} & \textbf{79.87} \\
    
        \bottomrule
        \end{tabular}
    }
    \end{center}
    % \vspace{-0.5em}
    \caption{\small 
    \textbf {F1 scores} in XNER on the datasets from CoNLL  and \cite{bari19}.  "--" represents no results were reported. %"x"  represents model fails to converge and
    %\newline
    %\emph{Fasttext-bi-LSTM+CRF} representes bi-directional LSTM model with a linear chain CRF and \emph{Char-bi-LSTM+Fasttext-bi-LSTM+CRF} means bi-directional hierarchical LSTM (character + word) with a linear chain CRF model. Both of the models are trained with mono-lingual Fasttext embeddings. "x"  represents model fails to converge and "-" represents no results are reported for the setup. \emph{BERT} and \emph{mBERT} represent monolingual and multilingual Language Model (LM) and \emph{cased} or \emph{uncased} represents if the LM is trained on case sensitive text or not. \emph{BERT-\textbf{base}} means 12 later LM was used for task adaptation. 
    }
    \label{table:ner1}
    % \vspace{-1em}
\end{table*}

%We transfer from English (en) to Spanish (es), German (de), Dutch (nl), Arabic (ar), Finnish (fi), Urdu (ur), Bengali (bn) and Burmese (my). 
%We collected the Arabic ( and Finnish datasets from \citet{bari19}. 

%For en and de, we consider the CoNLL'03 dataset \citep{Sang2003IntroductionTT}, while for es and nl, we use the CoNLL'02 dataset \citep{Sang2002IntroductionTT}.  
%For XNER, we transfer from English (en) to Spanish (es), German (de), Dutch (nl), Arabic (ar), and Finnish (fi). 

% \vspace{-0.5em}
\paragraph{XNLI} 
%XNLI judges the model's ability to extract a reasonable meaning representation of sentences across different languages. 
We use the standard dataset \citep{XNLI}. For a given pair of sentences, the  task is to predict the entailment relationship between the two sentences, \emph{i.e.}, whether the second sentence (\emph{hypothesis}) is an \textit{Entailment}, \emph{Contradiction}, or \textit{Neutral} with respect to the first one (\emph{premise}). We experiment with Spanish, German, Arabic, Swahili (sw), Hindi (hi) and Urdu.

\vspace{-0.5em}
\paragraph{PAWS-X} 
{The Paraphrase Adversaries from Word Scrambling Cross-lingual task \citep{yang-etal-2019-paws} requires the models to determine whether two sentences are paraphrases. We evaluate on all the six (typologically distinct) languages: fr, es, de, Chinese (zh), Japanese (ja), and Korean (ko).}

\paragraph{Evaluation setup}

Our goal is to adapt a task model from a source language distribution to an unknown target language distribution assuming no labeled data in the target. In this scenario, there might be two different distributional gaps: \Ni the generalization gap for the source distribution, and \Nii  the gap between the source and target language distribution. {We wish to investigate our method in tasks that exhibit such properties.} We use the standard task setting for XNER, where we take 100\% samples from the datasets as they come from various domains and sizes without any specific bias.

However, both XNLI and PAWS-X training data come with machine-translated texts in target languages. Thus, the data is parallel and lacks enough diversity (source and target come from the same domain). Cross-lingual models trained in this setup may pick up distributional bias (in the label space) from the source. \citet{artetxe2020translation} also argue that the translation process can induce subtle artifacts that may have a notable impact on models. 

%\vspace{-0.5em}
Therefore, for XNLI and PAWS-X, we experiment with two different setups. First, to ensure distributional differences and non-parallelism, we use 5\% of the training data from the source language and augment a different (nonparallel) 5\%  data for the target language. We used a different seed each time to retrieve this 5\% data.  Second, to compare with previous methods, we also evaluate on the standard 100\% setup. The evaluation is done on the entire test set in both setups. We will refer to these two settings as \textbf{5\%} and \textbf{100\%}. More details about model settings are in Appendix D.

%However, the XNLI dataset comes with machine-translated texts in target languages. 

%\blue{ We also test our method with the 100\% source (en) language dataset. }

%However, for XNLI, we use two different sizes of datasets.  and often contain dull and same domain data. 

%%%%%%%%%%%%%%%%%%%%%%%%%%%%%%%%%%%%%%%%%%
%%%%%%%%%%%%%%%%%%%%%%%%%%%%%%%%%%%%%%%%%%
%\red{Everything below in settings will go to supplementary}

% \vspace{-0.5em}
\subsection{Results}  
\label{sec:results}
%We present our results on XNER and XNLI in Tables \ref{table:ner} and  \ref{table:xnli}, respectively. Due to imbalance in class distribution, we use the predicted entity's F1 score for NER, which is the standard way of evaluating NER performance. For XNLI, we use the raw accuracy as there is no class imbalance in the dataset.

%For \textit{Single-model distillation} (\cref{subsec:distil}), we use \textit{distillation by model confidence} in all the presented results of \multimix\ in  Table \ref{table:ner} and \ref{table:xnli}. More on this in framework analysis (\cref{sec:analysis}) and  Appendi. %supplementary material.
%}

\vspace{-.5em}

\begin{table}
%\resizebox{0.5\columnwidth}{!}
\centering
\scalebox{0.7}{
    \begin{tabular}{lccc} 
    \toprule
    \bf{Model}  & \bf{ur} &\bf{bn} &\bf{my}\\
    \midrule
    \multicolumn{4}{c}{\textbf{Supervised Results}} \\
    \midrule
    XLM-R (our-impl) & 97.1 & 97.8 & 76.8\\
    \midrule
    \multicolumn{4}{c}{\textbf{Zero-Resource Results}} \\
    \midrule
    XLM-R ({XTREME}) & 56.4 & 78.8 & 54.3\\
    XLM-R ({our imp.}) & 56.45 & 78.17 & 54.56\\
    \multimix\ & \textbf{84.99} & \textbf{82.68} & \textbf{70.61}\\
    \bottomrule
    \end{tabular}
    }
    % \vspace{-0.5em}
    \caption{\small{XNER results on WikiANN}.}
    \label{ner:wikiann}
\end{table}

\begin{table*}[t]
% \vspace{-1.2em}
% \caption{Results in Accuracy for Cross-lingual Natural Language Inference (XNLI) task.}
\begin{center}
\resizebox{0.64\linewidth}{!}{
    \begin{tabular}{lccccccc}
    \toprule 
    \textbf{Model} & \textbf{en} & \textbf{es} & \textbf{de} & \textbf{ar} & \textbf{sw} & \textbf{hi} & \textbf{ur}\\
    % \midrule
    \midrule
    \multicolumn{8}{c}{\textbf{Supervised Results} (TRANSLATE-TRAIN-ALL)} \\
    % \midrule
    \midrule
    %XLM (Wiki) \cite{XLM}    & 84.5 & 81.3 & 79.3 & 75.6 & 69.2 & 72.1 & 67.7\\
    %XLM (Wiki+MT) \cite{XLM} & 85.0 & 80.3 & 80.3 & 77.6 & 72.8 & 72.9 & 68.5\\
    % Huang et al. (Wiki+MT) \cite{huang-etal-2019-unicoder} & 85.6 & 82.3 & 80.9 & 78.2 & 73.8 & 73.4 & 69.6\\
    % XLM-R (Base)  & 85.4 & 82.2 & 80.3 & 77.3 & 73.1 & 76.1 & 73.0\\
    {XLM-R } & 89.1 & 86.6 & 85.7 & 83.1 & 78.0 & 81.6 & 78.1 \\
    % XLM-R (our imp.)  & 88.2 & 85.3 & 84.2 & 81.9 & 77.2 & 80.7 & 76.8 \\
    % \midrule
    \midrule
    %\multicolumn{8}{c}{Supervision with \textbf{Full (100\%) English labeled} training set} \\
    \multicolumn{8}{c}{Zero-Resource Baseline for \textbf{Full (100\%) English labeled} training set} \\
    % \midrule
    \midrule
    %mBERT-uncased & 81.4 & 74.3 & 70.5 & 62.1 & - & - & 58.3 \\
    % \citet{artexXNLI} & 73.9 & 72.9 & 72.6 & 71.4 & 62.2 & 65.5 & 61.0\\
    %mBERT-cased \cite{xBERT} & 82.1 & 74.3 & 71.1 & 64.9 & 50.4 & 60.0 & 58.0 \\
    %XLM \cite{XLM} & 83.2 & 76.3 & 74.2 & 68.5 & 64.6 & 65.7 & 63.4\\

    %XLM-R (Paper) \cite{XLMR} & 89.1 & 85.1 & 83.9 & 79.8 & 73.9 & 76.9 & 73.8 \\
    XLM-R (XTREME) & 88.7 & 83.7 & 82.5 & 77.2 & 71.2 & 75.6 & 71.7 \\
    XLM-R (our imp.) & 88.87  & 84.34  & 82.78  & 78.44  & 72.08  & 76.40  & 72.10  \\
    
    XLM-R  (ensemble) & 89.24 & 84.73 & 83.27 & 79.06 & 73.17 & 77.23 &73.07 \\
    
    \midrule
    %\midrule
    % \multicolumn{8}{c}{Our Method} \\
    %\midrule
    % \midrule
    XLM-R+con-penalty & 88.83  & 84.30  & 82.86  & 78.20  & 71.83  & 76.24  & 71.62 \\
    
    \multimix\ & --  & {85.65 } & {84.15} & {80.50 } & {74.70 } & {78.74 } & 	{73.35 }\\
    
    \multimix\ (ensemble) & -- & \textbf{86.12} & \textbf{84.61} & \textbf{80.89} & \textbf{74.89} & \textbf{78.98} & \textbf{73.45} \\

    % \midrule
    \midrule   
    %\multicolumn{8}{c}{Supervision with \textbf{1\% English labeled} training set} \\
    \multicolumn{8}{c}{Zero-Resource Baseline for \textbf{5\% English labeled} training set} \\
    \midrule
    % \midrule
    
    XLM-R  (our imp.) & 83.08 & 78.48 & 77.54 & 72.04  & 67.3  & 70.41  & 66.72 \\
    
    XLM-R (ensemble) & 84.65 & 79.56 & 78.38 & 72.22 & 66.93 & 71.00 & 66.79 \\
    
    \midrule
    %\midrule
%   \multicolumn{8}{c}{Our Method} \\
    %\midrule
    % \midrule
    XLM-R+con-penalty & 84.24  & 79.23  & 78.47 & 72.43 & 67.72 & 71.08 & 67.63\\
    \multimix\ & -- & {81.53} & {80.88}  & {77.42} & {72.31} & {74.70} & {70.84}\\
    
    \multimix\ (ensemble) & -- & \textbf{82.35}   & \textbf{81.93} & \textbf{78.56} &\textbf{ 73.53} & \textbf{75.20}   & \textbf{71.15} \\
    \bottomrule
    \end{tabular}
}
\end{center}
% \vspace{-0.9em}
\caption{\small{Results in accuracy for XNLI.}}
\label{table:xnli1}
% \vspace{-1em}
\end{table*}

\paragraph{XNER}

%in addition to previous methods we also compare with a model \emph{ensemble} approach. %\red{We consider the average logit representation from different seed experiments in this purpose.}

% \begin{wraptable}{r}{6cm}
% %\vspace{-1.2em}

% \caption{\small{XNER results on WikiANN}}
%     %\resizebox{0.5\columnwidth}{!}
% %\vspace{-0.5em} 
% \centering
% \scalebox{0.7}{
%     \begin{tabular}{lccc} 
%     \toprule
%     \bf{Model}  & \bf{ur} &\bf{bn} &\bf{my}\\
%     \midrule
%     XLM-R \citep{XTREME} & 56.4 & 78.8 & 54.3\\
%     XLM-R ({Our imp.}) & 56.45 & 78.17 & 54.56\\
%     \multimix\ & \textbf{84.99} & \textbf{82.68} & \textbf{70.61}\\
%     \bottomrule
%     \end{tabular}
%     }
%     \label{ner:wikiann}
%     \vspace{-0.5em}
% \end{wraptable}

% \vspace{-.5em}
\Cref{table:ner1} reports the XNER results on the datasets from CoNLL  and \cite{bari19}, where we also evaluate an \emph{ensemble} by averaging the probabilities from the three models.
We observe that after  performing \textit{warm-up} with conf-penalty (\cref{subsec:conf-pen}), XLM-R performs better than mBERT on average by $\sim$3.8\% for all the languages. %On average, \multimix\ gives a sizable improvement of $\sim$5.5\% on five different languages. 
\multimix\ gives absolute improvements of 3.76\%, 4.34\%, 6.94\%, 8.31\%, and 4.18\% for \textit{es, nl, de, ar,} and \textit{fi}, respectively. Interestingly, it surpasses \emph{supervised}  {LSTM-CRF} for \textit{nl} and \textit{de} without using any target language labeled data. It also produces comparable results for \textit{es}. 

% Ensemble of base models (XLM-R) improves over the baseline scores by 0.90\% (on avg). \multimix\ still achieves sizeable improvements (4.6\%). The ensemble of \multimix\ models also gives an improvement of 0.41\% (on avg) over the base \multimix\.

In \Cref{ner:wikiann}, we report the results on the three \emph{low-resource} langauges from WikiANN. From these results and the results of \textit{ar} and \textit{fi} in \Cref{table:ner1}, we see that \multimix\ is particularly effective for languages that are structurally dissimilar and/or low-resourced, especially when the base model is weak: 28.54\%, 16.05\%, and 9.25\% absolute improvements for ur, my and ar, respectively.

\paragraph{XNLI-5\%}

From \Cref{table:xnli1}, we see that the performance of XLM-R trained on 5\% data is surprisingly good compared to the model trained on full data (see {XLM-R (our imp.)}), lagging by only {5.6\%} on average. In our single GPU implementation of XNLI, we could not reproduce the reported results of \citet{XLMR}. However, our results resemble the reported XLM-R results of XTREME \citep{XTREME}. We consider XTREME as our standard baseline for XNLI-100\%.

%the produced XLM-R results by \citet{XTREME} as the standard baseline for our XNLI experiments.
%As XNLI training datasets are parallel, for 5\% English labeled training set, we augment a different (not in the English) 5\% unlabeled dataset for the target language to disambiguate the parallel sense. 

%Our data augmentation method (\textbf{\multimix\-TopConf}) gives sizeable improvements over the baseline (mBERT-cased) across all the languages. Specifically, we get an improvement of 1.53\% for en, 1.2\% for es,	2.29\% for de,  3.09\% for ar, 0.84\% for sw, 3.55\% for hi, and 2.91\% for ur. 

%Table \ref{table:xnli} presents the results on the XNLI task. 
%We design our experiments in such a way that there will always be a \textit{distributional gap} between the train-test dataset distribution and the source-target language distribution.  

%To make the task more challenging and interesting, we augment only 1\% (randomly sampled) target training samples without the label. 
%From Table \ref{table:xnli}, we observe that 

%%%%%%%%%%%%%%%%%%%%%%%%%%%%%%%%%%%%%%%
%%%%%%%%%%%%%%%%%%%%%%%%%%%%%%%%%%%%%%%%

% \begin{table*}[h!]
% \caption{

We observe that with only 5\% labeled data in the source, \multimix\ gets comparable results to the {XTREME} baseline that uses 100\% labeled data (lagging behind by only $\sim$0.7\% on avg.); {even for \textit{ar} and \textit{sw}, we get 0.22\% and 1.11\% improvements, respectively}. It surpasses the standard 5\% baseline by 4.2\% on average. Specifically, \multimix\ gets absolute improvements of 3.05\%, 3.34\%, 5.38\%, 5.01\%, 4.29\%, and 4.12\% for \textit{es, de, ar, sw, hi,} and \textit{ur}, respectively. Again, the gains are relatively higher for low-resource and/or dissimilar languages despite the base model being weak in such cases.

% \vspace{-0.5em}
\paragraph{XNLI-100\%}

Now, considering \multimix's performance on the full (100\%) labeled source data in Table \ref{table:xnli1}, we see that it achieves SoTA results for all of the languages with an absolute improvement of 2.55\% on average from the {XTREME} baseline. Specifically, \multimix\ gets absolute improvements of 1.95\%, 1.68\%, 4.30\%, 3.50\%, 3.24\%, and 1.65\% for \textit{es, de, ar, sw, hi,} and \textit{ur}, respectively. %\red{We also see that the results of \multimix\ are better than the supervised results of XLM (Wiki) \cite{XLM} and Huang et al. (Wiki+MT) \cite{huang-etal-2019-unicoder}.}}

%%%%%%%%%%%%%%%%%%%%%%%%%%%%%%
%%%%%%%%%%%%%%%%%%%%%%%%%%%%%%%
%%%PAWSX
%%%%%%%%%%%%%%%%%%%%%%%%%%%%%%
%%%%%%%%%%%%%%%%%%%%%%%%%%%%%%%

\begin{table*}
\begin{center}
\resizebox{.65\linewidth}{!}{
    \begin{tabular}{lccccccc}
    \toprule 
    \textbf{Model} & \textbf{en} & \textbf{de} & \textbf{es} & \textbf{fr} & \textbf{ja} & \textbf{ko} & \textbf{zh}\\
    \midrule
    % \midrule

    \multicolumn{8}{c}{\textbf{Supervised Results} (TRANSLATE-TRAIN-ALL)} \\
    % \midrule
    \midrule
    {XLM-R (our impl.)} &  95.8 & 92.5 & 92.8 & 93.5 & 85.5 & 86.6 & 87.6 \\
    \midrule
    % \midrule

    \multicolumn{8}{c}{Zero-Resource Baseline for \textbf{Full (100\%) English labeled} training set} \\
    % \midrule
    \midrule
    XLM-R (XTREME) &  94.7 & 89.7 & 90.1 & 90.4 & 78.7 & 79.0 & 82.3 \\
    
    XLM-R (our imp.) & 95.46 & 90.06 & 89.92 & 90.85 & 79.89 & 79.74 & 82.49 \\
    
    XLM-R  (ensemble) & 96.10 & 90.75 & 90.55 & 91.80 & 80.55 & 80.70 & 83.45 \\
    
    \midrule
    XLM-R+con-penalty & 95.38 & 90.75 & 90.72 & 91.71 & 81.77 & 82.07 & 84.25 \\
    
    \multimix\ & --  & 92.27 & 92.28 & 92.85 & 83.88 & 84.27 & 86.90\\
    
    \multimix\ (ensemble) & -- & \textbf{92.55} &\textbf{ 92.35} & \textbf{93.35} & \textbf{84.30} & \textbf{84.35} & \textbf{86.95}\\

    % \midrule
    \midrule   
    \multicolumn{8}{c}{Zero-Resource Baseline for \textbf{5\% English labeled} training set} \\
    % \midrule
    \midrule
    
    XLM-R  (our imp.) & 91.15 & 83.72 & 84.32 & 85.08 & 73.65 & 72.60 & 77.22  \\
    
    XLM-R (ensemble) & 92.05 & 84.05 & 84.65 & 85.75 & 74.30 & 71.95 & 77.50\\
    
    \midrule
    XLM-R+con-penalty & 91.85 & 86.15 & 86.38 & 85.98 & 76.03 & 75.43 & 79.15 \\
    
    \multimix\ & --  & 89.05 & 90.27 & 90.12 & 80.50 & 79.60 & 82.65 \\
    
    \multimix\ (ensemble) & -- & \textbf{89.25} & \textbf{90.85} & \textbf{90.25} & \textbf{81.15} & \textbf{80.15} & \textbf{82.90} \\
    \bottomrule
    \end{tabular}}
\end{center}
% \vspace{-0.5em}
\caption{Results in accuracy for PAWS-X.}
\label{table:pawsx}
% \vspace{-1em}
\end{table*}

% \vspace{-0.5em}
\paragraph{PAWS-X}
Similar to XNLI, we observe sizable improvements for \multimix\ over the baselines on PAWS-X for both 5\% and 100\% settings (\Cref{table:pawsx}). Specifically, in 5\% setting, \multimix\ gets absolute gains of 5.33\%, 5.94\%, 5.04\%, 6.85\%, 7.00\%, and 5.45\% for \textit{de, es, fr, ja, ko,} and \textit{zh}, respectively, while in 100\% setting, it gets 2.21\%, 2.36\%, 2.00\%, 3.99\%, 4.53\%, and 4.41\% improvements respectively. In general, we get an average improvements of 5.94\% and 3.25\% in PAWS-X-5\% and PAWS-X-100\% settings respectively. Moreover, our 5\% setting outperforms 100\% XLM-R baselines for \textit{es, ja,} and \textit{zh}. Interestingly, in the 100\% setup, our \multimix\ (ensemble) achieves almost similar accuracies compared to supervised finetuning of XLM-R on all target language training dataset.

\section{Analysis}
\label{sec:analysis}

In this section, we analyze \multimix\ by dissecting it and measuring the contribution of its \emph{each of the components}. For this, we use the XNER task and analyze the model based on the results in Table \ref{table:ner1}.

%To understand which part of the framework provides the improvement, we perform a thorough analysis. More precisely, our goal here is to assess the contributions of \textit{data augmentation} and \textit{distillation} for our {self-learning} framework. 

% \vspace{-0.5em}
\subsection{Analysis of distillation methods}
%\vspace{-0.5em}

\paragraph{Model confidence vs. clustering} We first analyze the performance of our \textit{single-model distillation} methods (\cref{subsec:distil}) to see which of the two alternatives works better. From Table \ref{table:ner-distil}, we see that both perform similarly with \textit{model confidence} being slightly better. In our main experiments (Tables \ref{table:ner1}-\ref{table:pawsx}) and subsequent analysis, we use model confidence for distillation. However, we should not rule out the clustering method as it gives a more general solution to consider other distillation features (\eg\ sequence length, language) than model prediction scores, which we did not explore in this paper.  
%We further explored   and keep  for future exploration.

% \vspace{-0.5em}
\paragraph{Distillation factor $\eta$} We next show the results for different distillation factor ($\eta$) in Table \ref{table:ner-distil}. Here 100\% refers to the case when no single-model distillation is done based on model confidence. We notice that the best results for each of the  languages are obtained for values other than 100\%, which indicates that distillation is indeed an effective step in \multimix. See Appendix B for more analysis on $\eta$.   

% More on this hyperparameter ($\eta$) selection in the supplementary material.  
%We present the experimental results for \textit{distillation by model confidence} with different confidence values ($\eta$). 

%The results imply that our \textit{single model } is working. 

% \vspace{-0.5em}
\paragraph{Two-stage distillation}

We now validate whether the second-stage distillation (\textit{distillation by model agreement}) is needed. In \Cref{table:ner-distil}, we also compare the results with the model agreement (shown as $\cap$) to the results without using any agreement ($\phi$). We observe better performance with \textit{model agreement} in all the cases {on top of} the single-model distillation which validates its utility. Results with $\eta=100, Agreement=\cap$ can be considered as the tri-training \cite{tri-training} baseline.

\begin{table}
\centering
    \scalebox{0.66}{
    \begin{tabular}{ccccccc}
    \toprule 
    \textbf{$\eta$} & \textbf{Agreement}  & \textbf{es} & \textbf{nl} & \textbf{de} &\textbf{ar} & \textbf{fi}\\
    %$\eta$ & agreement & \\
    \midrule[0.9pt]    \multicolumn{7}{c}{Distillation by \textbf{clustering}}   \\
    \midrule[0.9pt]
  {0.7} & $\cap$ &  82.28   &  83.25   &  78.86   &  52.64  &  78.47     
    \\
    \midrule
    {0.5} & $\cap$ & 82.35  &  83.11   &  78.16   &  54.20  &  78.28 
    \\
    
    \midrule[0.9pt]
    \multicolumn{7}{c}{Distillation by  \textbf{model confidence}} \\
    \midrule[0.9pt]
    \multirow{ 2}{*}{50\%} & $\cap$ &  \textbf{82.52 } &  82.46  &  75.95   &  52.00   &  77.51  \\
    & $\phi$ &  81.66  &   82.26  &  77.19  &   52.97  &   77.77 \\
    \midrule
    
    \multirow{ 2}{*}{80\%}& $\cap$  &  82.33   &  \textbf{83.53 }  &  78.50 &  \textbf{54.48 }  &  78.43     
    \\
    & $\phi$ &  81.61 &   83.03 &   77.08 &   53.31 &   78.34\\
    \midrule
    
    \multirow{ 2}{*}{90\%}& $\cap$  &  81.90  &  82.80   &  \textbf{79.03}  &  52.41  &  \textbf{78.66 }  
    \\
    & $\phi$ &  81.21 & 82.77 & 77.28 & 52.20 & 77.93 \\
    \midrule
    
    \multirow{ 2}{*}{100\%} & $\cap$ &  82.50  &  82.35  &  77.06  &  52.58  &  77.51    
    \\
    & $\phi$ &  81.89 & 82.15 & 76.97  & 52.68 & 78.01 \\
    \bottomrule
    
    \end{tabular}
}
    % \vspace{-0.5em}
    \caption{\small Analysis of \textbf{distillation} on XNER. Results after epoch-1 training that uses $\{\gD_s, \gD'_t\}$.}\label{table:ner-distil}
    % \vspace{-.5em}
\end{table}

\begin{figure}
\centering
    \includegraphics[scale=.3]{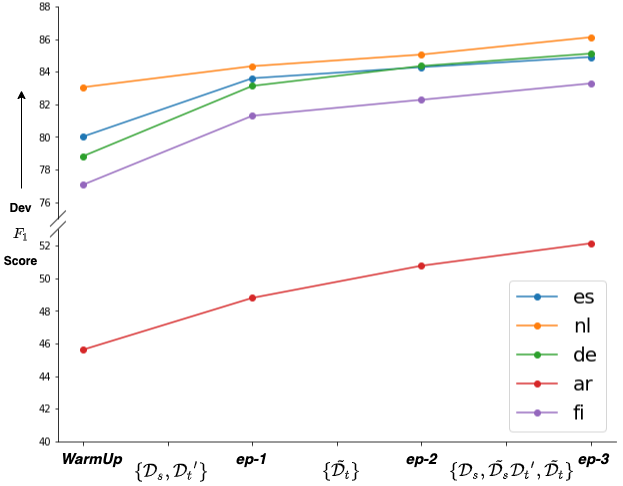}
    % \vspace{-0.5em}
    \caption{\small Validation F1 results in XNER for multi-epoch co-teaching training of \multimix.}
    \label{fig:multi-epoch}
\end{figure}

% \vspace{-0.5em}
\subsection{Augmentation in Stages}
%{\paragraph{Benefits of different types of augmentation (mix) in different stages} 
Figure \ref{fig:multi-epoch} presents the effect of different types of augmented data used {by different epochs} in our multi-epoch co-teaching framework. We observe that in every epoch, there is a significant boost in F1 scores for each of the languages. Arabic, being structural dissimilar to English, has a lower base score, {but the relative improvements brought by \multimix\ are higher for Arabic, especially in epoch 2 when it gets exposed to the target language virtual data ($\Tilde{\gD}_t$) generated by the vicinity distribution.}

%\red{Put a comment on comparison with tri-training (without vicinal augmentation?)}

%and distilled pseudo labels.  

%than other languages 

%MultiMix turns out to be even more effective in such cases yielding more relative gains.  

%\red{Compared to other languages, zero-shot transfer for Arabic is low. We suspect this is because of the structural dissimilarity between Arabic and English \cite{k2020crosslingual}. However, the improvement for Arabic is consistent with (and even higher than) other languages.}

% table:ner1, wikiann, xnli1, pawsx

% \vspace{-0.5em}
\subsection{Effect of Confidence Penalty \& Ensemble}

For all the three tasks, we get reasonable improvements over the baselines by training with confidence penalty (\cref{subsec:conf-pen}). Specifically, we get 0.56\%, 0.74\%, 1.89\%, and 1.18\% improvements in XNER, XNLI-5\%, PAWS-X-5\%, and PAWS-X-100\% respectively (Table \ref{table:ner1},\ref{table:xnli1},\ref{table:pawsx}). The improvements in XNLI-100\% are marginal and inconsistent, which we suspect due to the balanced class distribution. %Details visualization of confidence penalty is on \Cref{app:conf}.

%Interestingly, we do not get improvements for XNLI-100\% which we suspect due to the balance label space of full XNLI training data. 

From the results of ensemble models, we see that the ensemble boosts the baseline XLM-R. However, our regular \multimix\ still outperforms the ensemble baselines by a sizeable margin. Moreover, ensembling the trained models from \multimix\ further improves the performance. These comparisons ensure that the capability of \multimix\ through co-teaching and co-distillation is beyond the ensemble effect.

% \vspace{-0.5em}
\subsection{Robustness \& Efficiency}

Table \ref{table:robustness} shows the robustness of the fine-tuned \multimix\  model on XNER task. After fine-tuning in a specific target language, the F1 scores in English remain almost similar (see first row). For some languages, \multimix\ adaptation on a different language also improves the performance. For example, Arabic gets improvements for all \multimix-adapted models (compare 50.88 with others in row 5). This indicates that augmentation of \multimix\ does not overfit on a target language. \textbf{{More baselines, analysis and visualizations are added in Appendix.}} 

% \red{Put a comment on why using 3 models does not loose efficiency}

\begin{table}[t]
\centering
\scalebox{0.7}{
    \begin{tabular}{ccccccc}
    \toprule 
    \textbf{Tgt} & Zero shot + & \multicolumn{5}{c}{\textbf{\multimix}}   \\
     \textbf{lang} & con-penalty  &  \textbf{es} & \textbf{nl} & \textbf{de} & \textbf{ar} & \textbf{fi}\\
    
    \midrule
    en  & \textbf{92.88}  & \textbf{92.92}  & \textbf{92.87}   & \textbf{92.91}  & \textbf{92.80}  & \textbf{92.68}  \\ %
    es  & 81.42  & \textbf{83.24}  & 82.01   & 77.71  & 80.29  & 81.97  \\ %
    nl  & 81.27  & 81.22  & \textbf{85.32}   & 80.54  & 82.36  & 84.20  \\ %
    de  & 75.20  & 73.63  & 75.03   & \textbf{80.03}  & 76.97  & 73.77  \\ %
    ar  & 50.88  & 52.66  & 53.08   & 52.52  & \textbf{58.29}  & 53.80  \\ %
    fi  & 76.97  & 77.02  & 77.06   & 76.69  & 77.13  & \textbf{80.11}  \\ %
    \bottomrule
    \end{tabular}
}
% \vspace{-0.5em}
\caption{\small {\textbf{F1 scores} on XNER. Each  column (\eg\ es) under \textbf{\multimix} represents results in all target languages for a \multimix\ trained with the augmented data in a specific language (\eg\ es). The \textbf{Zero shot+con-penalty} column represents the zero-shot results for the model after \textbf{WarmUp}.}}
\label{table:robustness}
\end{table}

%\red{We compare our MultiMix frmaework with the XLM-R ensemble baselines.  From Table \ref{table:ner1},\ref{table:xnli1},\ref{table:pawsx} we see that ensemble baselines of XLM-R models always surpass the regular baselines. }

%{if we take the  of  from MultiMix framework, we get even  better scores in each of the tasks which implies that capability of MultiMix framework is beyond the ensemble effect.} 

%We consider the average logit representation from different seed experiments in this purpose.

% \red{We are not including the baseline "jointly training with a target language modeling" objective}

% \vspace{-0.5em}
\section{Related Work}  \label{sec:background}

Recent years have witnessed significant progress in learning multilingual pretrained models. %\citep{howard-ruder-2018-universal,ELMO,BERT,XLNet,radford2019language}. 
Notably, mBERT \citep{BERT} extends (English) BERT by jointly training on 102 languages. XLM \citep{XLM} extends mBERT with a conditional LM and a translation LM (using parallel data) objectives. \citet{XLMR} train the largest multilingual language model XLM-R with RoBERTa \citep{RoBERTa}. \citet{xBERT}, \citet{Keung_2019}, and \citet{pires-etal-2019-multilingual} evaluate zero-shot cross-lingual transferability of mBERT  on several tasks and attribute its generalization capability
to shared subword units. \citet{pires-etal-2019-multilingual} also found structural similarity (\eg\ word order) to be another important factor for successful cross-lingual transfer. \citet{k2020crosslingual}, however, show that the shared subword has a minimal contribution; instead, the structural similarity between languages is more crucial for effective transfer. 

Older data augmentation approaches relied on distributional clusters \cite{tackstrom2012cross}. 
A number of recent methods have been proposed using contextualized LMs  \citep{kobayashi,conditional_bert,Shi2019AUGBERTAE,bosheng-et-al-emnlp-20,linlin-et-al-acl-21}. These methods rely on labels to perform label-constrained augmentation, thus not directly comparable with ours. Also, there are fundamental differences in the way we use the pretrained LM. Unlike them our LM  augmentation is purely unsupervised and we do not perform any fine-tuning of the pretrained vicinity model. This disjoint characteristic gives our framework the flexibility to replace $\theta_{lm}$ even with a better monolingual LM for a specific target language, which in turn makes \multimix\ extendable to utilize stronger LMs that may come in the future. {In a concurrent work \cite{mohiuddin-et-al-augvic}, we propose a contextualized LM based data augmentation for neural machine translation and show its advantages over traditional back-translation gaining improved performance in low-resource scenarios.}

\section{Conclusion}
% \vspace{-0.5em}

We propose a novel data augmentation framework, \multimix, for zero-resource cross-lingual task adaptation. It performs simultaneous self-training with data augmentation and unsupervised sample selection. With extensive experiments on three different cross-lingual tasks spanning many language pairs, we have demonstrated the effectiveness of \multimix. For the zero-resource XNER task, \multimix\ sets a new SoTA for all the {tested languages}. For both XNLI and PAWS-X tasks, with only 5\% labeled data in the source, \multimix\ gets comparable results to the baseline that uses 100\% labeled data. 
%Through an in-depth analysis, we show the cumulative contributions of different components of \multimix.

\bibliographystyle{imports/acl_natbib.bst}
\bibliography{imports/ref.bib}

%\appendix
\clearpage

\appendix
\section*{Appendix}

\section{FAQ: Justifications for design methodology of \multimix}
\label{app:design_choice}
 \label{app:justification}

Here are our justifications for various design principles of the \multimix\ framework.

\paragraph{Is masked language model pre-training with cross-lingual training data from task dataset useful?}

In Table \ref{table:base}, We perform language model finetuning on XLM-R large model with multilingual sentences of NER dataset and perform adaptation with only English language. With the LM-finetuned XLM-R model, we didn't see any significant increase in cross-lingual transfer. For Spanish, Arabic language, the score even got decreased, which indicates possible over-fitting. However, robustness experiment in table 6 (see in the main paper, sec 4.4) indicates that our proposed method doesn't overfit on target language rather than augment the new knowledge base.

\begin{table}[h!]
    \centering 
    \resizebox{1\columnwidth}{!}
    {
        \begin{tabular}{lccccc} 
            \toprule
            Model  & es & nl & de & ar & fi\\
            \midrule
            XLM-R  & 80.45	& 81.07	& 73.76	& 49.94	& 76.05 \\
            XLM-R + ens  & 81.42  & 	81.27 & 	75.20 & 	50.93 &	76.97\\
            \multimix\ & 83.05 &	85.21 &	80.33 &	57.35 &	79.75\\
            \multimix\ + ens & 83.24 & 	85.32 & 	80.99 & 	58.29 & 	79.87\\
            \midrule
            Finetuned XLM-R & 78.11 & 	81.61 & 	76.33 & 	48.04 & 	76.63\\
            \bottomrule
        \end{tabular}
    }
    \caption{Some additional baseline results on XNER task. Here, ens reefers to emsemble.}
    \label{table:base}
\end{table}

%\vspace{-0.5em}
\paragraph{Is using three models with different initialization necessary? }
Yes, different initialization ensures different convergence paths, which results in diversity during inference. Co-labeling (Section 3.3) utilizes this property. There could be some other ways to achieve the same thing. Our initial attempt with three different heads (sharing a backbone network) didn't work well.

\paragraph{Is using three epochs necessary? }
We utilize different types of datasets in different epochs. While pseudo-labeling may induce noise, the model's predictions for in-domain cross-lingual samples are usually better. Because of this, for a smooth transition, we apply the vicinal samples in the second epoch. Finally, inspired by the joint training of the cross-lingual language model,  in the third epoch we use all four datasets. We also include the labeled source data which ensures that our model does not overfit on target distribution as well as persists the generalization capability of the source distribution.

\paragraph{Need for the combination of co-teaching, co-distillation and co-guessing?}
The combination of these helps to distill out the noisy samples better.

\paragraph{Efficiency of the method and expensive extra costs for large-scale pretrained models}
It is a common practice in model selection to train 3-5 disjoint LM-based task models (e.g., XLM-R on NER) with different random seeds and report the ensemble score or score of the best (validation set) model. In contrast, \multimix\ uses 3 different models and jointly trains them where the models assist each other through distillation and co-labeling. In that sense, the extra cost comes from distillation and co-labeling, which is not significant and is compensated by the significant improvements that \multimix\ offers.

\section{Visualization of confidence penalty} 
\label{app:conf}
\subsection{Effect of confidence penalty in classification} \label{app:penalty-class}

\begin{figure*}

%\label{table:img-para_all}
\begin{minipage}[t]{0.23\textwidth}
    \centering
    \includegraphics[scale=.18]{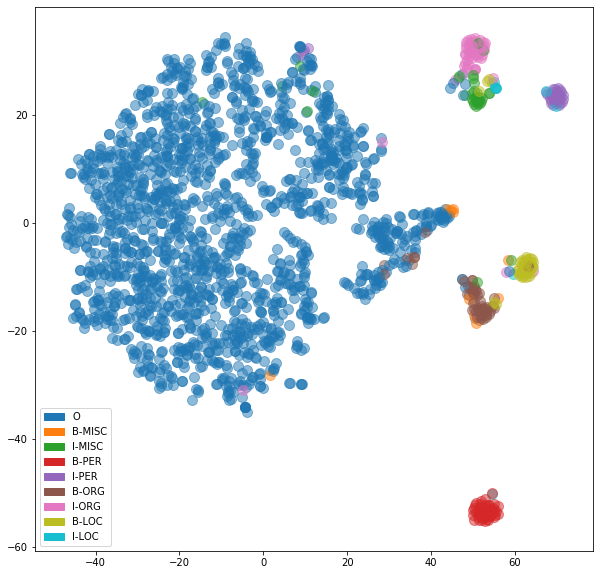}
    \caption*{(a) Without confidence penalty.}
\end{minipage}
\hfill
\begin{minipage}[t]{0.23\textwidth}
    \centering
    \includegraphics[scale=.18]{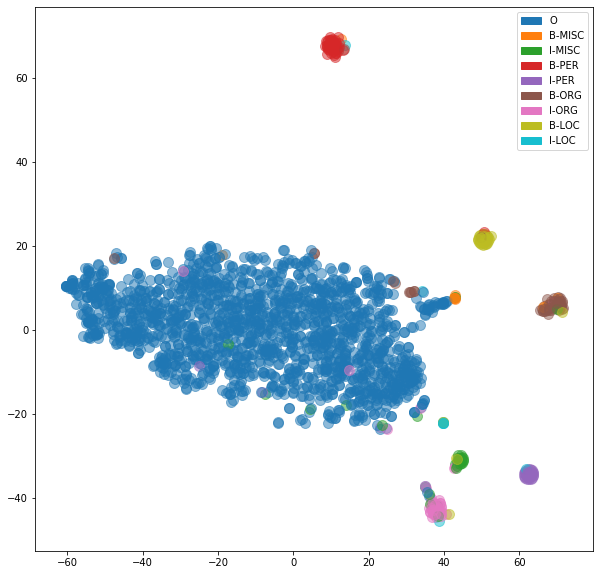}
    \caption*{(b) With confidence penalty.}
\end{minipage}
\hfill
\begin{minipage}[t]{0.23\textwidth}
    \centering
    \includegraphics[scale=.19]{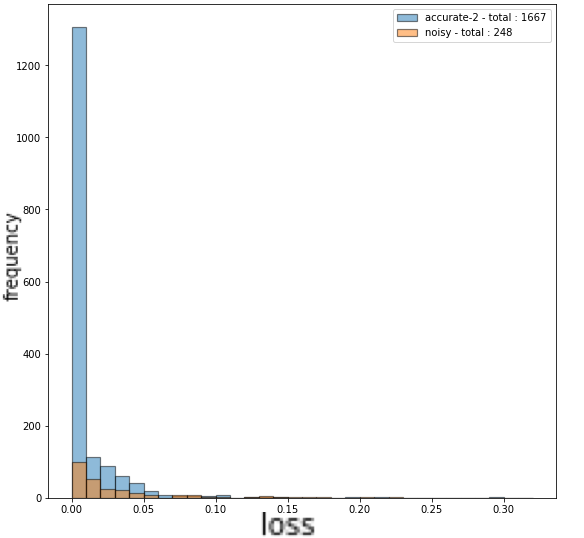}
    \caption*{(c) Without confidence penalty.}
\end{minipage}
\hfill
\begin{minipage}[t]{0.23\textwidth}
    \centering
    \includegraphics[scale=.19]{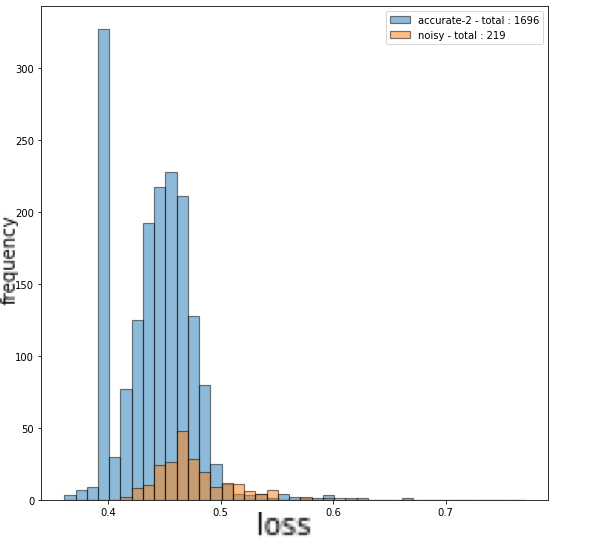}
    \caption*{(d) With confidence penalty.}
\end{minipage}

\caption{(a-b) Effect of training with confidence penalty in the warm-up step on target (\textit{Spanish}) language XNER classification using t-SNE plots. From the visualization, it can be seen that the model trained with confidence penalty shows better inter-class separation which exhibits robustness of the multilingual model. (c-d) Histogram of loss distribution on target (\textit{Spanish}) language XNER classification.}
\label{fig:conf-penalty}
\end{figure*}

In Figure \ref{fig:conf-penalty} (a-b), we present the effect of the confidence penalty {(Eq. 1 in the main paper)} in the target language (\textit{Spanish}) classification on the XNER dev. data {(\ie\ after training on English NER)}. We show the class distribution from the final logits (on the target language)  using t-SNE plots. From the figure, it is evident that the use of confidence penalty in the warm-up step makes the model more robust to unseen out-of-distribution target language data yielding better predictions, which in turn also provides a better \emph{prior} for self-training with pseudo labels. 

%\clearpage
\subsection{Effect of confidence penalty in loss distribution}
\label{apendix:c-2}

Figures \ref{fig:conf-penalty}(c) and \ref{fig:conf-penalty}(d) present the per-sample loss (\ie\ mean loss per sentence \wrt\ the pseudo labels) distribution in histogram without and with confidence penalty, respectively. Here, \textit{accurate-2} refers to the sentences which have at most two wrong NER labels, and sentences containing more than two errors are referred to as  \textit{noisy} samples. {It shows that without confidence penalty, there are many noisy samples with a small loss which is not desired.} In addition to that, the figures also suggest that the confidence penalty helps to separate the clean samples from the noisy ones either by clustering or by model confidence.

\begin{figure*}[h!]
%\label{table:img-para_all}
\begin{minipage}[t]{0.24\textwidth}
    \centering
    \includegraphics[scale=.2]{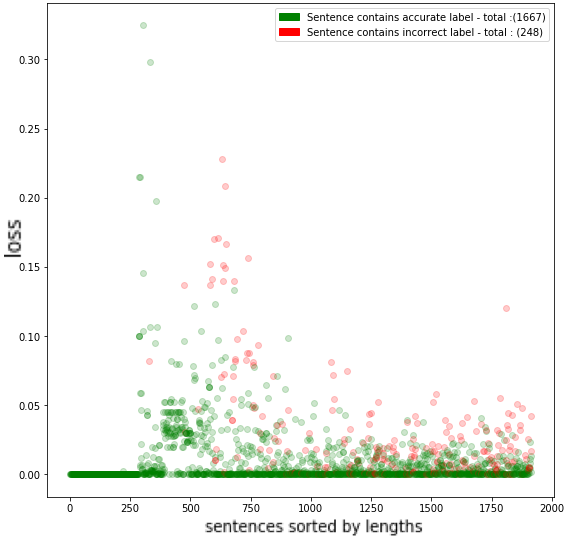}
    \caption*{(a) Without confidence penalty.}
\end{minipage}
\hfill
\begin{minipage}[t]{0.24\textwidth}
    \centering
    \includegraphics[scale=.2]{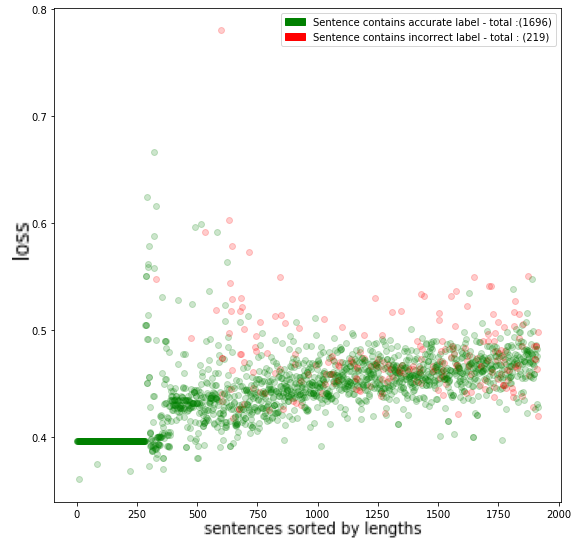}
    \caption*{(b) With confidence penalty.}
\end{minipage}
\hfill
\begin{minipage}[t]{0.24\textwidth}
    \centering
    \includegraphics[scale=.205]{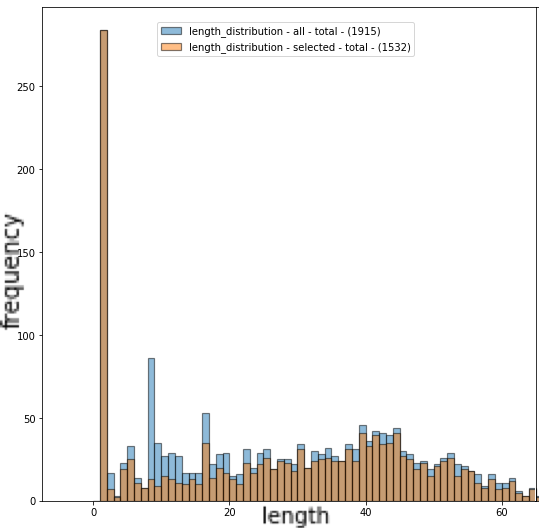}
    \caption*{(a) Without confidence penalty.}
\end{minipage}
\hfill
\begin{minipage}[t]{0.24\textwidth}
    \centering
    \includegraphics[scale=.205]{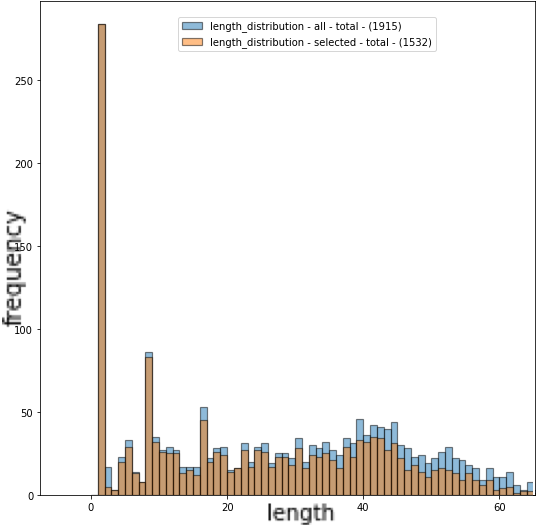}
    \caption*{(b) With confidence penalty.}
\end{minipage}

\caption{(a-b)Scatter plot of loss distribution on target (\textit{Spanish}) language XNER classification. (c-d) Distribution of selected sentence lengths on target (\textit{Spanish}) language XNER classification.}
\label{fig:loss-scatter}
\end{figure*}

Figures \ref{fig:loss-scatter}(a) and \ref{fig:loss-scatter}(b) present the loss distribution in a scatter plot by sorting the sentences based on their length in the x-axis; y-axis represents the loss. As we can see, the losses are indeed more scattered when we train the model with confidence penalty, which indicates higher per-sample entropy, as expected. Also, we can see that as the sentence length increases, there are more wrong predictions. Our distillation method should be able to distill out these noisy pseudo samples. 

Finally, Figures \ref{fig:loss-scatter}(c) and \ref{fig:loss-scatter}(d) show the length distribution of all vs. the selected sentences (by \emph{Distillation by model confidence}) without and with confidence penalty. Bari et al. (2020) shows that cross-lingual NER inference is heavily dependent on the length distribution of the samples. In general, the performance of the lower length samples is more accurate. However, if we only select the lower length samples we will easily overfit. From these plots, we observe that the confidence penalty also helps to perform a better distillation as more sentences are selected (by the distillation procedure) from the lower length distribution, while still covering the entire lengths. This shows that using the confidence penalty in training, model becomes more robust.

In summary, comparing the Figures \ref{fig:conf-penalty}(c-d) - \ref{fig:loss-scatter}(c-d), we can conclude that training without confidence penalty can make the model more prone to over-fitting, resulting in more noisy pseudo labels. Training with confidence penalty not only improves pseudo labeling accuracy but also helps the distillation methods to perform better noise filtering.

\section{Details on distillation by clustering} \label{app:clustering}
One limitation of the confidence-based (single-model) distillation is that it does not consider task-specific information. Apart from classifier confidence, there could be other important features that can distinguish a good sample from a noisy one. For example, for sequence labeling, \emph{sequence length} can be an important feature as the models tend to make more mistakes (hence noisy) for longer sequences Bari et al. (2020). One might also want to consider other features like \emph{fluency}, which can be estimated by a pre-trained conditional LM like GPT Radford et al. (2020). In the following, we introduce a clustering-based method that can consider these additional features to separate good samples from bad ones. 

Here our goal is to cluster the samples based on their \emph{goodness}. It has been shown in computer vision that deep models tend to learn good samples faster than noisy ones, leading to a lower loss for good samples and higher loss for noisy ones Han et al. (2018), Arpit et al. (2017).  We propose to model \textit{per-sample loss distribution} (along with other task-specific features) with a mixture model, which we fit using an \textit{Expectation-Maximization} (EM) algorithm. However, contrary to those approaches which use actual (supervised) labels, we use the model predicted pseudo labels to compute the loss for the samples. 

We use a two-component Gaussian Mixture Model (GMM) due to its flexibility in modeling the sharpness of a distribution Li et al. (2020a). In the following, we describe the EM training of the GMM for one feature, \ie\ per-sample loss, but it is trivial to extend it to consider other indicative task-specific features like sequence length or fluency score (see any textbook on machine learning).   

\paragraph{EM training for two-component GMM}

Let $x_i \in \real$ denote the loss for sample $\mathbf{x}_i$ and $z_i \in \{0, 1\}$ denote its cluster id. We can write the 1d GMM model as:   
\begin{equation}
p(x_i|\theta,\pi) = \sum_{k=0}^1 \gN (x_i|\mu_k, \sigma_k) \pi_k    
\end{equation}
where $\theta_k = \{ \mu_k, \sigma^2_k \}$ are the parameters of the $k$-th mixture component and $\pi_k = p(z_i = k)$ is the probability (weight) of the $k$-th component with the condition $0 \le \pi_k \le 1$ and $\sum_k \pi_k = 1$.  

In EM, we optimize the \emph{expected complete data} log likelihood $Q(\theta, \theta^{t-1})$ defined as:

\begin{align}
\begin{aligned}
&Q(\theta, \theta^{t-1})\\ 
&=  \E( \sum_i \log [ p (x_i, z_i|\theta)]) \\
&=  \E( \sum_i \sum_k \mathbb{I}(z_i = k) \log [ p (x_i|\theta_k) \pi_k ]) \\
&=  \sum_i \sum_k \E (\mathbb{I}(z_i = k)) \log [ p (x_i|\theta_k) \pi_k ] \\
&=  \sum_i \sum_k p (z_i = k|x_i, \theta^{t-1}) \log [ p (x_i|\theta_k) \pi_k ] \\ 
&=  \sum_i \sum_k r_{i,k} (\theta^{t-1}) \log p (x_i|\theta_k)  + r_{i,k} (\theta^{t-1}) \log  \pi_k \label{eq:em}
\end{aligned}
\end{align}

where $r_{i,k} (\theta^{t-1})$ is the responsibility that cluster $k$ takes for sample $\mathbf{x}_i$, which is computed in the E-step so that we can optimize $Q(\theta, \theta^{t-1})$ (Eq. \ref{eq:em}) in the M-step. The E-step and M-step for a 1d GMM can be written as: 

\textbf{E-step: } Compute $r_{i,k} (\theta^{t-1}) = \frac{ \gN (x_i|\theta_k^{t-1}) \pi_k^{t-1}}{\sum_k \gN(x_i|\theta_k^{t-1}) \pi_k^{t-1}}$

\textbf{M-step:} Optimize $Q (\theta,\theta^{t-1})$ \wrt\  $\theta$ and $\pi$

\begin{itemize}

\item $\pi_k = \frac{\sum_i r_{i,k}}{\sum_i \sum_k r_{i,k}} = \frac{1}{N} {\sum_i  r_{i,k}}$

\item $\mu_k = \frac{\sum_i r_{i,k} x_i}{ \sum_i r_{i,k}}$; \hspace{2em} $\sigma^2_k = \frac{\sum_i r_{i,k} (x_i - \mu_k)^2}{\sum_i r_{i,k}}$

\end{itemize}

\paragraph{Inference}

For a sample $\mathbf{x}$, its \emph{goodness} probability is the posterior probability $p(z=g|\mathbf{x}, \theta)$, where $g \in \{0,1\}$ is the component with smaller mean loss. Here, distillation hyperparameter $\eta$ is the posterior probability threshold based on which samples are selected.

% \red{the following can be moved to appendix if we need space}

\paragraph{Relation with \emph{distillation by model confidence}}
Astute readers might have already noticed that per-sample loss has a direct deterministic relation with the model confidence. Even though they are different, these two distillation methods consider the same source of information. However, as mentioned, the clustering-based method allows us to incorporate other indicative features like length, fluency, etc. For a fair comparison between the two methods, we use only the per-sample loss in our primary (single-model) distillation methods.

%in the clustering model in supervised set-up. 

\section{Hyperparameters} \label{app:settings}
\begin{table*}[!]
\centering
\resizebox{.9\linewidth}{!}{
    \begin{tabular}{lcccccc}
    \toprule 
    \textbf{Hyperparameter} & \multicolumn{2}{c}{\textbf{XNER}} & \multicolumn{2}{c}{\textbf{XNLI}} & \multicolumn{2}{c}{\textbf{PAWS-X}}\\
    \cmidrule(lr){2-3}\cmidrule(lr){4-5} \cmidrule(lr){6-7}
    & \textbf{Warm-up step} & \textbf{X-lingual adaptation} & \textbf{Warm-up step} & \textbf{X-lingual adaptation} & \textbf{Warm-up step} & \textbf{X-lingual adaptation}\\
    
    \midrule
    \midrule
    \multicolumn{7}{c}{\textbf{Training-hyperparameters}} \\
    \midrule
    \midrule
    model-type & \texttt{xlm-r L} & \texttt{warm-up-ckpt} & \texttt{xlm-r L} & \texttt{warm-up-ckpt} & \texttt{xlm-r L} & \texttt{warm-up-ckpt}\\
    sampling-factor $\alpha$ & -- & 0.7  & -- & 0.7 & -- & 0.7\\
    drop-out  & 0.1  & 0.1  & 0.1  & 0.1& 0.1  & 0.1\\
    max-seq-length  & 280 & 280 & 128 & 128& 128 & 128\\
    per-gpu-train-batch-size & 4 & 4 & 16 & 16& 16 & 16 \\
    {grad-accumulation-steps} & 5  & 4  & 2  & 2& 2  & 2 \\
    logging-step & 50 & 50  & 50 & 25 & 50 & 25 \\
    learning-rate (lr) & 3$e^{-5}$ & 5$e^{-6}$ & 1$e^{-6}$ & 1$e^{-6}$ & 1$e^{-6}$ & 1$e^{-6}$\\
    {lr-warm-up-steps}  & 200 & 10\% of train  & 10\% of train & 10\% of train& 10\% of train & 10\% of train\\
    weight-decay & 0.01  & 0.01 & -- & --& -- & -- \\
    adam-epsilon & 1$e^{-8}$ & 1$e^{-8}$ & 1$e^{-8}$ & 1$e^{-8}$& 1$e^{-8}$ & 1$e^{-8}$\\
    max-grad-norm & 1.0  & 1.0 & 1.0  & 1.0 & 1.0  & 1.0 \\
    num-of-train-epochs & -- & 1  & -- & 1 & -- & 1 \\
    \multimix-epochs & -- & 3 & 6 & 3 & 10 & 6 \\
    max-steps & 3000 & --  & -- & -- & -- \\
    train-data-percentage & 100 & 100 & 5 & 5 & 5 & 5\\
    conf-penalty & \texttt{True}  & \texttt{False}  & \texttt{True}  & \texttt{False} & \texttt{True}  & \texttt{False} \\

    \midrule
    \midrule
    \multicolumn{7}{c}{\textbf{Distillation-hyperparameters}} \\
    \midrule
    \midrule   
    \#mixture-component & -- &. 2 & -- & --& -- & --\\
    posterior-threshold & -- & 0.5 & -- & -- & -- & -- \\
    covariance-type & -- & \texttt{Full} & -- & -- & -- & --\\
    distilation-factor $\eta$ & -- & 80, 100, 100 & -- & 50, 80, 100 & -- & 80, 90, 80\\
    distillation-type & -- & \texttt{confidence} & -- & \texttt{confidence} & -- & \texttt{confidence}\\

    \midrule
    \midrule
    \multicolumn{7}{c}{\textbf{Augmentation-hyperparameters}} \\
    \midrule
    \midrule
    do-lower-case & \texttt{False} & \texttt{False} & \texttt{False} & \texttt{False} & - & \texttt{False}\\
    aug-type & -- & \texttt{successive-max} & -- & \texttt{successive-cross} & -- & \texttt{successive-cross}\\
    aug-percentage $P$ & -- & 30  & -- & 30 &  -- &  40\\
    diversification-factor $\delta$ & -- & 3 & -- & 2$\times$2 & -- & 2 $\times$ 2\\

    \bottomrule
    \end{tabular}
}
\caption{Hyperparameter settings for XNER, XNLI, and PAWS-X task. Total number of parameter for each of the model is 550M. We used V100 GPUs to do the experiments. Average run-time for each of the languages may differ based on total number of augmented samples. In an average, for per million augmentation requires ~.5-2 days based of various settings of training mechanism (ie., fp16 training, gradient accumulation etc).}
\label{table:hyp}
\end{table*}

We present the hyperparameter settings for XNER and XNLI tasks for the XLA framework in Table \ref{table:hyp}. {In the \textit{warm-up} step, we train and validate the task models with English data. However, for \textit{cross-lingual adaptation}, we validate (for model selection) our model with the target language development set.} We train our model with respect to the number of steps instead of the number of epochs. In the case of a given number of epochs, we convert it to a total number of steps.

 \noindent We observe that \textit{learning rate} is a  crucial hyperparameter. In table \ref{table:hyp}, \textit{lr-warm-up-steps} refer to the \emph{warmup-step} from triangular learning rate scheduling. This hyperparameter is not to be confused with \textit{Warm-up step} of the \multimix\ framework. In our experiments, effective \textit{batch-size} is another crucial hyperparameter that can be obtained  by gradient accumulation steps. We fix the maximum sequence length to $280$ for XNER and $128$ {tokens} for XNLI.

\noindent For each of the experiments, we report the average score of three task models, $\theta^{(1)}$, $\theta^{(2)}$, $\theta^{(3)}$, which are initialized with different seeds. We perform each of the experiments in a single GPU setup with \textit{float32} precision.

\section{Additional Related Work} \label{app:ext-rel-work}

\paragraph{Contextual representation and cross-lingual transfer.}

In earlier approaches, word representations are learned from simple variants of the skip-gram model \cite{mikolov2013distributed}, where each word has a single representation regardless of its context \cite{grave2018learning,GLOVE}. Recent approaches learn word representations that change based on the context that the word appears in \cite{mccann2017learned,ELMO,howard-ruder-2018-universal,BERT,XLNet,radford2019language}. 

\citet{ELMO} propose {ELMo} - a bidirectional LSTM-based LM pre-training method for learning contextualized word representations. ELMo uses a linear combination of all of its layers' representations for predicting on a target task. However, because of sequential encoding, LSTM-based LM pre-training is hard to train at scale. \citet{vaswani_trans} propose the Transformer architecture based on multi-headed self-attention and positional encoding. The Transformer encoder can capture long-range sequential information and  allows {constant time} encoding of a sequence through parallelization. {\citet{Radford2018ImprovingLU}} propose {GPT-1}, which pre-trains a Transformer decoder with a conditional language model objective and then fine-tune it on the task with minimal changes to the model architecture. In the same spirit, \citet{BERT} propose {BERT}, which pre-trains a Transformer encoder with a masked language model (MLM) objective, and uses the same model architecture to adapt to a new task. The advantage of MLM objective is that it allows bidirectional encoding, whereas the standard (conditional) LM is unidirectional (\emph{i.e.}, uses either left context or right context).

BERT also comes with a multilingual version called mBERT, which has  12 layers, 12 heads and 768 hidden dimensions, and it is trained jointly on 102 languages with a shared vocabulary of 110K subword tokens.\footnote{\href{github.com/google-research/bert/blob/master/multilingual.md}{github.com/google-research/bert/blob/master/multilingual.md}} Despite any explicit cross-lingual supervision, mBERT has been shown to learn cross-lingual representations that generalise well across languages. \citet{xBERT,pires-etal-2019-multilingual} evaluate the zero-shot cross-lingual transferability of mBERT  on several NLP tasks and attribute its generalization capability to shared subword units. \citet{pires-etal-2019-multilingual} additionally found structural similarity (\emph{e.g.},\ word order) to be another important factor for successful cross-lingual transfer. \citet{k2020crosslingual}, however, show that the shared subword has minimal contribution, rather the structural similarity between languages is more crucial for effective transfer. \citet{artetxe2019crosslingual} further show that joint training may not be necessary and propose an alternative method to transfer a monolingual model to a {bi-lingual model} through learning only the word embeddings  in the target language. They also identify the vocabulary size per language as an important factor.

\citet{XLM} extend mBERT with a conditional LM and a translation LM (using parallel data) objectives and a language embedding layer. They train a larger model with more monolingual data. \citet{huang-etal-2019-unicoder} propose to use auxiliary tasks such as cross-lingual word recovery and paraphrase detection for pre-training.  Recently, \citet{XLMR} train the largest multilingual language model with 24-layer transformer encoder, 1024 hidden dimensions and 550M parameters. \citet{Keung_2019} use adversarial fine-tuning of mBERT to achieve better language invariant contextual representation for {cross-lingual} NER and {MLDoc} {document classification}.

%\vspace{-1em}
\paragraph{Vicinal risk minimization.}

One of the fundamental challenges in deep learning is to train models that generalize well to examples outside the training distribution. The widely used Empirical Risk Minimization (ERM) principle where models are trained to minimize the average training error has been shown to be insufficient to achieve generalization on distributions that differ slightly from the training data \cite{Szegedy2014,zhang2018mixup}. Data augmentation supported by the Vicinal Risk Minimization (VRM) principle \cite{VRM_NIPS2000} can be an effective choice for achieving better out-of-training generalization.

In VRM, we minimize the empirical vicinal risk defined as:

\vspace{-1em}
\begin{equation}
    \gL_v (\theta) = \frac{1}{N} \sum_{n=1}^{N} l(f_{\theta}( \Tilde{x}_n ), \Tilde{y}_n) 
\end{equation}
\vspace{-1em}

\noindent where $f_{\theta}$ denotes the model parameterized by $\theta$, and $\gD^{\text{aug}} = \{ (\Tilde{x}_n, \Tilde{y}_n) \}_{n=1}^{N}$ is an augmented dataset constructed by sampling the vicinal distribution $\vartheta(\Tilde{x}, \Tilde{y}|{x}_i, {y}_i)$ around the original training sample $({x}_i, {y}_i)$. Defining vicinity is however challenging as it requires to extract samples from a distribution without hurting the labels. Earlier methods apply simple rules like rotation and scaling of images \cite{Simard1998}. Recently, \citet{zhang2018mixup,MixMatch} and \citet{Li2020DivideMix} show impressive results in image classification with simple linear interpolation of data. However, to our knowledge, none of these methods has so far been successful in NLP due to the discrete nature of texts.

\end{document}